\documentclass[a4paper]{cas-sc}

\usepackage[numbers]{natbib}
\usepackage{array}
\usepackage{amsmath}
\usepackage{booktabs}
\usepackage{multirow}
\usepackage{tikz}
\usetikzlibrary{positioning,calc,shapes.geometric}
\newcolumntype{C}[1]{>{\centering\arraybackslash}p{#1}}

\begin{document}
\let\WriteBookmarks\relax
%% Float placement: pack floats onto text pages and require float pages to be full.
\renewcommand{\topfraction}{0.92}
\renewcommand{\bottomfraction}{0.75}
\renewcommand{\textfraction}{0.06}
\renewcommand{\floatpagefraction}{0.85}
\setcounter{topnumber}{3}
\setcounter{bottomnumber}{2}
\setcounter{totalnumber}{5}
%% On float-only pages, stack the floats from the top instead of spreading
%% them over the full height, so any slack collects at the foot of the page.
\makeatletter
\setlength{\@fptop}{0pt}
\setlength{\@fpsep}{10pt}
\setlength{\@fpbot}{0pt plus 1fil}
\makeatother

\shorttitle{ECG Monitoring under Energy Constraints}
\shortauthors{M. F. R. Ibrahim et~al.}

\title[mode = title]{Robust Remote ECG Monitoring under Energy Constraints: Anomaly Filtering versus Predictive Uncertainty}
%\tnotetext[1]{This article is a substantially extended version of the authors' paper presented at the PerconAI workshop of the 2026 IEEE International Conference on Pervasive Computing and Communications~\cite{ibrahim2026enhancing}. Section~\ref{subsec:contributions} states what is new.}

\author[1,3,4]{Mustafa Fuad Rifet Ibrahim}[orcid=0009-0008-3575-5688]
\cormark[1]
\ead{mustafa.ibrahim@nxp.com}
\credit{Conceptualization, Data curation, Formal analysis, Investigation, Methodology, Software, Validation, Visualization, Writing -- Original Draft, Writing -- Review \& Editing}

\affiliation[1]{organization={CTO System Innovation, NXP Semiconductors Germany GmbH},
	addressline={Beiersdorfstraße 12}, 
	city={Hamburg},
	postcode={22529},
	country={Germany}}

\author[2]{Maurice Meijer}
\credit{Writing -- Review \& Editing, Supervision}

\affiliation[2]{organization={Business Line Secure Connected Edge, NXP Semiconductors},
	addressline={High Tech Campus 60}, 
	city={Eindhoven},
	postcode={5656},
	country={The Netherlands}}

\author[3]{Alexander Schlaefer}
\credit{Writing -- Review \& Editing, Supervision}

\affiliation[3]{organization={Institute of Medical Technology and Intelligent Systems, Hamburg University of Technology},
	addressline={Am Schwarzenberg-Campus 1},
	city={Hamburg},
	postcode={21073},
	country={Germany}}

\author[4]{Peer Stelldinger}
\credit{Writing -- Review \& Editing, Supervision}

\affiliation[4]{organization={Faculty of Computer Science and Digital Society, Hamburg University of Applied Sciences},
	addressline={Berliner Tor 5},
	city={Hamburg},
	postcode={20099},
	country={Germany}}

\cortext[cor1]{Corresponding author}

\begin{abstract}
Machine learning for remote electrocardiogram monitoring must handle noise and diseases absent from training under tight resource constraints. We investigate whether upstream unsupervised anomaly detection can improve monitoring decisions more efficiently than classifier-derived uncertainty. We evaluate five detectors on PTB-XL and BUT QDB and compare a two-stage filter--classifier system with Monte Carlo Dropout and deep ensembles under unseen-disease and noise conditions. Performance is averaged over five training seeds. We measure single-pass energy, latency, and graph memory on a microcontroller's central processing unit (CPU) and neural processing unit (NPU). Per-score neural network costs account for repeated evaluations but exclude scoring outside deployed graphs. Energy equations relate filtering and rejection to avoided classifier execution. Across four disease scenarios, Deep Support Vector Data Description (Deep SVDD) achieves the highest mean area under the receiver operating characteristic curve within all tested finite CPU energy budgets. On the NPU, it leads at 1\,mJ, while Monte Carlo Dropout has slightly higher means at 10 and 100\,mJ. With a narrow CPU classifier, the Deep SVDD two-stage system increases the fraction of correctly classified or correctly rejected inputs by 15.0--19.4 percentage points over classification without rejection. Estimated energy per input is 1.93--6.99\,mJ, compared with 52.53\,mJ for Monte Carlo Dropout and 8.76--17.51\,mJ for deep ensembles, at each method's retrospectively selected test accuracy peak. These results support upstream filtering for efficient remote monitoring, while independent validation and complete-system measurements remain necessary.
\end{abstract}


\maketitle

%% ===========================================================================
\section{Introduction}
\label{sect:intro}

Deploying machine learning models for continuous remote electrocardiogram (ECG) monitoring on wearable devices requires reliable predictions under severe energy, memory, and computational constraints~\cite{ibrahim2026ondevice}. Two challenges are particularly relevant to prediction reliability. Physiological noise, including electrode-motion artifacts, can degrade ECG classification performance~\cite{venton2021robustness}. Separately, a deployed classifier may encounter disease classes absent from its training data, creating an out-of-distribution (OOD) detection problem~\cite{barandas2024evaluation}. A classifier with a fixed set of diagnostic outputs cannot explicitly identify an omitted disease class, even when the recording has adequate signal quality. Moreover, neural networks can make confident errors under distribution shift, so confidence in a predicted label does not establish that the input is suitable for classification~\cite{ovadia2019can}.

One strategy for limiting unreliable predictions is to assess incoming recordings and withhold classification when an input is judged unsuitable. Studies of automated ECG interpretation have shown that uncertainty-based rejection can improve classification performance among retained recordings~\cite{barandas2024evaluation}. An independent unsupervised anomaly detection (UAD) filter offers a complementary approach: it learns from the intended input distribution and assesses each recording before the diagnostic classifier is invoked~\cite{ibrahim2026enhancing}. In this setting, in-distribution data include healthy ECGs and the disease classes represented during training. Anomaly detection therefore concerns deviations from this distribution, rather than the presence of disease alone. Rejecting a recording withholds the downstream prediction. It does not determine that the recording lacks clinical relevance.

Our previous work investigated this upstream filtering architecture using six UAD approaches: Deep Support Vector Data Description (Deep SVDD), autoencoders (AEs), variational autoencoders (VAEs), masked anomaly detection (MAD), normalizing flows (NFs), and denoising diffusion probabilistic models (DDPMs)~\cite{ibrahim2026enhancing}. We optimized these approaches through neural architecture search under a limit of 512,000 parameters and found that Deep SVDD provided a favorable balance between detection performance and parameter count. However, that study assessed efficiency through parameter count, leaving the hardware costs of filtering and its comparison with established uncertainty quantification (UQ) methods unresolved. An upstream filter adds computation for every recording but avoids classifier execution for rejected inputs. Its energy benefit therefore depends on the filter cost, the classifier cost, and the rejection rate. Detection performance and model size alone do not resolve this trade-off.

In this work, we extend the comparison to Monte Carlo Dropout (MC-Dropout)~\cite{gal2016dropout} and deep ensembles (DE)~\cite{lakshminarayanan2017simple}, two established approaches to predictive UQ. MC-Dropout uses repeated stochastic evaluations of a classifier, whereas DE aggregate predictions from independently trained classifiers. Including these methods allows us to evaluate the utility of independent upstream UAD filters against uncertainty estimates obtained from the diagnostic models themselves. We measure energy consumption and latency per forward pass through each successfully deployed model graph, together with static random-access memory (SRAM) usage. These measurements provide estimates of neural network execution costs per score after accounting for repeated forward passes. We also formulate energy equations relating filter and classifier costs to the rejection rate. By accounting for filtering every input and classifying only accepted inputs, these equations identify when upstream UAD is more or less energy efficient than MC-Dropout or DE.

\section{Related Work}
\label{sect:related_work}

ECG anomaly detection provides the methodological basis for our upstream filters. One established use is to learn the characteristics of healthy recordings and identify disease-related deviations. Atamny et~al.~\cite{atamny2023outlier} compared AEs, VAEs, NFs, diffusion models, and Gaussian mixture models for this purpose. Zhou and Kan~\cite{zhou2021tensor} applied Deep SVDD to features extracted from multilead ECG tensors to detect deviations from normal cardiac activity. Within this line of work, reconstruction and forecasting methods use errors in restoring or predicting ECG signals to identify abnormalities. Jiang et~al.~\cite{jiang2023multi} combined whole-record and heartbeat-level restoration to detect and localize abnormalities, while Ruiz-Barroso et~al.~\cite{ruiz2025fade} used forecasting errors relative to learned normal ECG dynamics to identify abnormal beats and rhythms. Computational cost has also been examined for reconstruction-based detectors. Bui et~al.~\cite{bui2024tsrnet} combined time-series and spectrogram representations for ECG restoration and evaluated inference latency, while Zhou et~al.~\cite{zhou2026lightweight} developed a multiscale masked AE with reduced parameter and operation counts. Anomaly detection can also support a diagnostic pipeline through learned representations or by selecting beats for further classification. Jiang et~al.~\cite{jiang2024self} used self-supervised anomaly-detection pretraining to improve recognition of rare diagnostic classes. Wang et~al.~\cite{wang2023hierarchical} first detected abnormal beats using an adversarially trained, memory-augmented AE and then classified those beats into arrhythmia types. Our filter instead treats healthy ECGs and represented disease classes as admissible inputs. It aims to reject recordings that belong to unseen disease classes or have insufficient signal quality before a separate classifier runs. Known pathological recordings remain candidates for classification.

Assessing whether a recording is suitable for a particular diagnostic model connects our work to signal-quality assessment and the detection of unknown disease classes. The first concern is whether noise has obscured the information needed for ECG analysis. Satija et~al.~\cite{satija2017automated} used signal decomposition, temporal features, and decision rules to detect, localize, and categorize ECG noise. Learning-based quality assessment overlaps more directly with the UAD methods considered here. Huang et~al. augmented Deep SVDD with synthetic-noise pseudo anomalies for ECG quality assessment~\cite{huang2024pseudo} and subsequently developed a joint time-frequency hypersphere model~\cite{huang2025joint}. Fan et~al.~\cite{fan2026global} combined AE residuals with global signal features for quality prediction. Han et~al.~\cite{han2025diffusion} used diffusion reconstruction to quantify noise and examined its relationship to downstream arrhythmia classification. Such assessments can also guide rejection before diagnosis: Rahman et~al.~\cite{rahman2025impact} placed a noise-detection and rejection stage before ECG analysis, assessed classification performance, and implemented the system on a smartphone. A separate concern is whether a well-recorded ECG contains evidence of a disease class absent from training. Vranken et~al.~\cite{vranken2021uncertainty} and Barandas et~al.~\cite{barandas2024evaluation} examined rejection of such recordings using classifier-derived uncertainty, including dropout and ensemble methods. Elul et~al.~\cite{elul2021meeting} combined multiple binary rhythm classifiers with MC-Dropout and an indicator of unfamiliar datasets derived from learned features. Yu et~al.~\cite{yu2025trustworthy} used logit-based Energy scores and ReAct activation truncation to detect unknown arrhythmias. A different response to unseen classes is the UIRD framework of Shi et~al.~\cite{shi2025unsupervised}, which combines generative novelty detection with replay-based learning to incorporate new ECG classes. Execution cost is a further consideration when applying these approaches to wearable devices. Rahman et~al.~\cite{rahman2026uncertainty} benchmarked UQ for Holter-based atrial fibrillation detection under noise and external validation, reporting inference time, parameter counts, and operation counts. Jia et~al.~\cite{jia2024ur2m,jia2026cascade} evaluated evidential classifiers with uncertainty-guided early exits on microcontrollers, including ECG experiments, and reported measured inference energy and latency together with memory usage. Our extension compares independently trained upstream UAD filters with MC-Dropout and DE for both unusable signals and held-out disease classes. Unlike uncertainty obtained through diagnostic-model evaluations, the upstream score can be computed without executing the classifier. For successfully deployed models, we measure energy and latency per forward pass and SRAM usage. We estimate neural network costs per score by accounting for repeated passes. Energy equations then relate these estimated costs to the classifier evaluations avoided at different rejection rates.

\section{Methodology}
\label{sect:methodology}

We study ECG monitoring on wearable devices with limited energy, memory, and computational resources. Our objective is to prevent a diagnostic classifier from issuing predictions for recordings affected by excessive noise or disease classes absent from training. We compare two ways of making this rejection decision and examine their detection performance and execution costs.

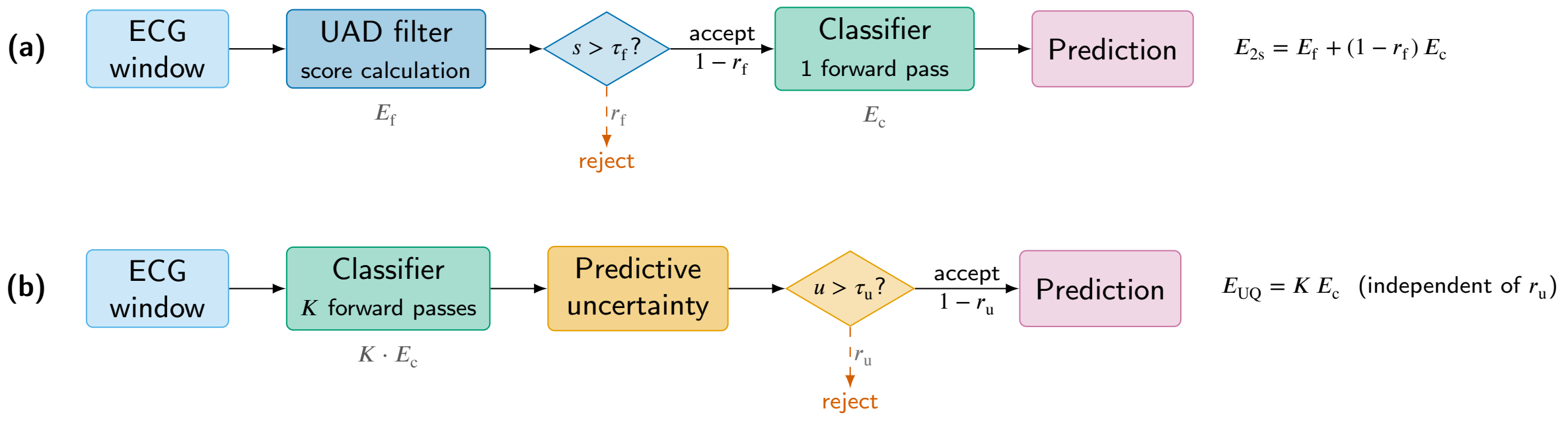
\begin{figure}[pos=t]
  \centering
  \resizebox{\linewidth}{!}{% Two-stage UAD filter vs. classifier-derived uncertainty quantification.
% Included from the main document; requires tikz + the listed libraries.
% Okabe--Ito colors distinguish the processing stages in both paths.
\definecolor{oiBlue}{HTML}{0072B2}
\definecolor{oiSkyBlue}{HTML}{56B4E9}
\definecolor{oiGreen}{HTML}{009E73}
\definecolor{oiOrange}{HTML}{E69F00}
\definecolor{oiPurple}{HTML}{CC79A7}
\definecolor{oiVermillion}{HTML}{D55E00}
\begin{tikzpicture}[
  font=\small,
  >=latex,
  box/.style   = {draw, rounded corners=2pt, minimum height=8mm, align=center,
                  inner xsep=4pt, inner ysep=3pt},
  sens/.style  = {box, draw=oiSkyBlue, fill=oiSkyBlue!30, minimum width=15mm},
  filt/.style  = {box, draw=oiBlue, fill=oiBlue!35, minimum width=21mm},
  clf/.style   = {box, draw=oiGreen, fill=oiGreen!35, minimum width=21mm},
  unc/.style   = {box, draw=oiOrange, fill=oiOrange!45, minimum width=19mm},
  term/.style = {box, draw=oiPurple, fill=oiPurple!30, minimum width=17mm},
  lbl/.style   = {font=\scriptsize, inner sep=1pt},
  cost/.style  = {font=\scriptsize\itshape, inner sep=1pt, text=black!65},
  drop/.style  = {->, dashed, oiVermillion},
]

% ------------------------------------------------------------------ row A
\node[sens] (aIn) at (0,0) {ECG\\window};
\node[filt, right=6mm of aIn] (aF) {UAD filter\\\scriptsize score calculation};
\node[diamond, draw=oiBlue, aspect=1.7, inner sep=1pt, fill=oiBlue!20,
      right=6mm of aF, font=\scriptsize] (aD) {$s>\tau_{\mathrm{f}}$?};
\node[clf,  right=11mm of aD] (aC) {Classifier\\\scriptsize 1 forward pass};
\node[term,  right=6mm of aC] (aO) {Prediction};
\node[lbl, below=6.5mm of aD, text=oiVermillion] (aR) {reject};

\draw[->] (aIn) -- (aF);
\draw[->] (aF)  -- (aD);
\draw[->] (aD)  -- node[lbl, above] {accept} node[lbl, below]{$1-r_{\mathrm f}$} (aC);
\draw[->] (aC)  -- (aO);
\draw[drop] (aD) -- node[lbl, right, text=black!55]{$r_{\mathrm f}$} (aR);

\node[cost, below=1.2mm of aF]  {$E_{\mathrm{f}}$};
\node[cost, below=1.2mm of aC]  {$E_{\mathrm{c}}$};
\node[anchor=west, font=\scriptsize] at ($(aO.east)+(3mm,0)$)
      {$E_{2\mathrm{s}}=E_{\mathrm{f}}+(1-r_{\mathrm f})\,E_{\mathrm{c}}$};
\node[anchor=east, font=\small\bfseries] at ($(aIn.west)+(-3mm,0)$) {(a)};

% ------------------------------------------------------------------ row B
\node[sens, below=17mm of aIn] (bIn) {ECG\\window};
\node[clf, right=6mm of bIn, minimum width=21mm] (bC)
      {Classifier\\\scriptsize $K$ forward passes};
\node[unc, right=6mm of bC] (bU) {Predictive\\uncertainty};
\node[diamond, draw=oiOrange, aspect=1.7, inner sep=1pt, fill=oiOrange!30,
      right=6mm of bU, font=\scriptsize] (bD) {$u>\tau_{\mathrm{u}}$?};
\node[term, right=11mm of bD] (bO) {Prediction};
\node[lbl, below=6.5mm of bD, text=oiVermillion] (bR) {reject};

\draw[->] (bIn) -- (bC);
\draw[->] (bC)  -- (bU);
\draw[->] (bU)  -- (bD);
\draw[->] (bD)  -- node[lbl, above] {accept} node[lbl, below]{$1-r_{\mathrm u}$} (bO);
\draw[drop] (bD) -- node[lbl, right, text=black!55]{$r_{\mathrm u}$} (bR);

\node[cost, below=1.2mm of bC] {$K\cdot E_{\mathrm{c}}$};
\node[anchor=west, font=\scriptsize] at ($(bO.east)+(3mm,0)$)
      {$E_{\mathrm{UQ}}=K\,E_{\mathrm{c}}$\ \ (independent of $r_{\mathrm u}$)};
\node[anchor=east, font=\small\bfseries] at ($(bIn.west)+(-3mm,0)$) {(b)};

\end{tikzpicture}}
  \caption{Upstream filtering and classifier-derived uncertainty. (a) The UAD filter rejects a fraction $r_{\mathrm f}$ of ECG windows at threshold $\tau_{\mathrm f}$, avoiding classifier execution for those inputs. (b) UQ-based rejection at threshold $\tau_{\mathrm u}$ follows all classifier evaluations. Here, $K$ denotes the number of MC-Dropout passes. For DE, the member count $M$ replaces $K$. UQ energy is estimated from one measured classifier pass multiplied by $K$ or $M$. All energy estimates exclude score calculations outside the deployed graphs. The equations assume equal classifier costs and exclude escalation.}
  \label{fig:system}
\end{figure}

Figure~\ref{fig:system} illustrates the two processing paths. An independent UAD filter assigns an anomaly score to each ECG window before classification. The classifier runs only if the score does not exceed a rejection threshold. The second path uses MC-Dropout or DE to derive uncertainty from multiple classifier evaluations. These evaluations provide both the prediction and the uncertainty score, so they must be completed before an input can be rejected.

Let $E_{\mathrm f}$ denote the estimated energy of the UAD model evaluations needed for one score, $E_{\mathrm c}>0$ the energy of one classifier forward pass, and $r_{\mathrm f}$ the filter's rejection rate. The expected inference energy per input window is
\begin{equation}
  E_{2\mathrm s}=E_{\mathrm f}+(1-r_{\mathrm f})E_{\mathrm c}.
  \label{eq:filter_energy}
\end{equation}
The filter cost is incurred for every input, whereas the classification cost applies only to accepted inputs. For UQ, we measure the energy of one forward pass through the underlying classifier and multiply it by the number of evaluations. Under the equal-cost approximation, the estimated energy for MC-Dropout is
\begin{equation}
  E_{\mathrm{UQ}}=K E_{\mathrm c},
  \label{eq:uq_energy}
\end{equation}
independently of the UQ rejection rate $r_{\mathrm u}$. For DE with $M$ members, $M$ replaces $K$. Equation~\eqref{eq:uq_energy} assumes that each evaluation costs $E_{\mathrm c}$. For all methods, the estimates exclude score calculations outside the deployed graphs, as detailed in Section~\ref{subsec:measurement}. Sensing and other shared operations are excluded from both paths.

To express the comparison independently of the absolute energy scale, define $\rho=E_{\mathrm f}/E_{\mathrm c}$. Dividing the condition $E_{2\mathrm s}<E_{\mathrm{UQ}}$ by $E_{\mathrm c}$ gives
\begin{equation}
  \rho+1-r_{\mathrm f}<K
  \quad\Longleftrightarrow\quad
  \rho<K-1+r_{\mathrm f}.
  \label{eq:energy_break_even}
\end{equation}
The right-hand side specifies how costly the filter can be, relative to one classifier evaluation, while retaining an energy advantage over UQ. More UQ evaluations increase this allowance. A higher filter rejection rate also increases it by reducing the downstream classification cost.

Rejection may itself incur an additional cost, for example when a recording is transmitted for further assessment. Let $E_{\mathrm{esc}}$ be the same energy cost per rejected input in either path. The expected costs then become
\begin{align}
  E_{2\mathrm s,\mathrm{esc}}&=E_{\mathrm f}+(1-r_{\mathrm f})E_{\mathrm c}+r_{\mathrm f}E_{\mathrm{esc}},\nonumber\\*
  E_{\mathrm{UQ},\mathrm{esc}}&=K E_{\mathrm c}+r_{\mathrm u}E_{\mathrm{esc}}.
  \label{eq:escalation_energy}
\end{align}
The upstream filter uses less energy when
\begin{equation}
  \rho<K-1+r_{\mathrm f}
  +(r_{\mathrm u}-r_{\mathrm f})\frac{E_{\mathrm{esc}}}{E_{\mathrm c}}.
  \label{eq:escalation_break_even}
\end{equation}
Equal rejection rates cancel the escalation term. If the filter rejects more inputs than UQ, escalation reduces its energy advantage. Energy savings are assessed alongside detection and classification performance because rejecting useful recordings also changes system utility.

\subsection{Datasets and preprocessing}
\label{subsec:datasets}

We retain the datasets and general evaluation design of our previous work~\cite{ibrahim2026enhancing}. The datasets are available through PhysioNet~\cite{pollard2026physionet,goldberger2000physionet}. PTB-XL, version 1.0.3, provides 21,799 ten-second, 12-lead ECG recordings from 18,869 patients~\cite{wagner2022ptbxl,wagner2020ptbxl}. We use the 500\,Hz recordings and five diagnostic superclasses: normal ECG (NORM), myocardial infarction (MI), conduction disturbance (CD), ST/T change (STTC), and hypertrophy (HYP). Labels are not mutually exclusive. We follow the recommended patient-respecting folds: folds 1--8 for training, fold 9 for validation, and fold 10 for testing. In each of four OOD scenarios, one disease superclass is withheld. All training records carrying that label are excluded, including records that also carry an in-distribution (ID) label. Validation and test records retain both ID and OOD cases.

Experiment~1 evaluates unseen-disease detection on these PTB-XL partitions and signal-quality rejection separately on the Brno University of Technology ECG Quality Database (BUT QDB)~\cite{nemcova2020butqdb}. BUT QDB contains 18 long-term, single-lead recordings sampled at 1,000\,Hz during daily activities. Its three quality classes distinguish signals with identifiable ECG morphology, signals with reliably identifiable QRS complexes but reduced morphological detail, and signals unsuitable for analysis. We treat classes 1 and 2 as admissible inputs and class 3 as excessive noise. We separate patients between training and evaluation and label a ten-second window anomalous when at least one second is annotated as class 3. This standalone noise task evaluates UAD methods only. With class 3 withheld, pooling classes 1 and 2 leaves a single training category. Consequently, the MC-Dropout and DE baselines have no nontrivial classification task.

Experiment~2 evaluates the complete rejection-and-classification paths on PTB-XL with both unseen diseases and added noise. The MIT-BIH Noise Stress Test Database supplies two-channel electrode-motion noise~\cite{moody1984noise}. We implement PhysioNet's QRS-amplitude signal-to-noise ratio (SNR) convention in Python~\cite{moody2022nst}. Noise severity is sampled from an empirical prior estimated from BUT QDB class-3 windows using QRS-based signal scaling and wavelet denoising. We apply db6 denoising at level 11 independently to each 10,000-sample window, with universal soft thresholding~\cite{donoho1995denoising}. We resample the noise excerpts to 500\,Hz and reuse the two noise channels cyclically across the 12 leads, scaling them separately for each lead. Corrupted intervals have durations sampled uniformly from 1 to 10\,s and randomly chosen feasible start times. Boundary offsets suppress discontinuities where the added noise begins and ends. Approximately 16.5\% of the recordings in each ID and OOD validation and test partition are modified, matching the class-3 window prevalence in the BUT QDB test set. This produces recordings with known diagnostic labels and controlled corruption, including cases with both noise and an unseen disease.

All model inputs represent ten-second windows resampled to 512 time steps. PTB-XL inputs use z-score normalization, while BUT QDB inputs use instance normalization. Noise is added before this final resampling and normalization. Sections~4 and~5 detail the protocols and evaluation metrics for Experiments~1 and~2, respectively.

\subsection{UAD and UQ methods}
\label{subsec:methods}

The six UAD methods learn from admissible inputs without examples of the anomalies to be detected. Deep SVDD~\cite{ruff2018deep} uses a one-dimensional residual network to map these inputs near a fixed hypersphere center. Its anomaly score is the squared distance from that center. AEs~\cite{hinton2006reducing} and VAEs~\cite{kingma2013auto} use convolutional encoders and decoders to reconstruct the input. Both methods use the $L_2$ reconstruction error as the anomaly score. The VAE also regularizes its latent distribution. MAD~\cite{fu2022mad} trains a transformer to reconstruct inputs with 5\% of time steps replaced by random values. At inference, reconstruction errors are accumulated over 512 evaluations, masking each time step in turn. NFs use a multiscale Glow architecture with affine coupling and invertible $1\times1$ convolutions~\cite{kingma2018glow}. We use negative log-likelihood as the anomaly score. DDPMs~\cite{ho2020denoising} use a one-dimensional U-Net to reconstruct partially noised inputs through iterative denoising, following the anomaly-detection approach of Wyatt et~al.~\cite{wyatt2022anoddpm}. Reconstruction error provides the score. Thus, a complete MAD or DDPM score can require multiple network evaluations.

For each UAD method, we retain the random architecture search used in our earlier work: 100 trials with at most 512,000 parameters per model~\cite{ibrahim2026enhancing}. For Deep SVDD, AE, and VAE, the search varies the number of layers and filters, convolutional kernel sizes and strides, and latent dimensions. For MAD, it varies transformer size, attention heads, patch size, and patch overlap. It also varies the Kullback--Leibler loss weight for VAE, coupling-subnetwork and flow structure for NF, and U-Net architecture, noising and denoising steps, and prediction targets for DDPM. The complete search spaces and implementation will be provided in the code released upon acceptance. The parameter limit constrains the architecture search. We measure energy, latency, and SRAM requirements after deployment.

The downstream classifier and both UQ methods use the \texttt{resnet1d\_wang} architecture, whose performance on PTB-XL diagnostic superclasses was established by Strodthoff et~al.~\cite{strodthoff2020deep}. For both methods, we sweep network-width multipliers of $0.25$, $0.5$, and $1.0$. MC-Dropout retains stochastic dropout during inference~\cite{gal2016dropout}, with $K\in\{10,20,30\}$ evaluations and dropout probabilities of $0.3$, $0.4$, or $0.5$. We place dropout throughout feature extraction rather than only before the classification head, allowing stochastic variation in intermediate as well as final representations~\cite{gal2016bayesianconv}. DE aggregate independently initialized and trained classifiers~\cite{lakshminarayanan2017simple}, with $M\in\{3,5,10\}$ members.

Each classifier outputs one sigmoid probability per label. For either $T=K$ stochastic evaluations or $T=M$ DE members, let $p_{t\ell}$ be the probability assigned to label $\ell$ by evaluation $t$, and let $\bar p_\ell=T^{-1}\sum_{t=1}^{T}p_{t\ell}$. We score epistemic uncertainty by summing labelwise mutual-information estimates~\cite{houlsby2011bayesian}:
\begin{equation}
  u(x)=\sum_{\ell=1}^{L}\left[
  h(\bar p_\ell)-\frac{1}{T}\sum_{t=1}^{T}h(p_{t\ell})\right],
  \qquad h(p)=-p\log p-(1-p)\log(1-p),
  \label{eq:uq_score}
\end{equation}
where $L$ is the number of predicted labels. Predictive entropy $h(\bar p_\ell)$ measures total uncertainty. Mean conditional entropy estimates its aleatoric component. Their difference measures disagreement attributable to the sampled models. We use the sum of these marginal estimates, not the mutual information of the joint label vector. Our preliminary tests favored this score over summed predictive entropy for anomaly detection. We train the classifiers for supervised classification, following the original UQ formulations~\cite{gal2016dropout,lakshminarayanan2017simple}. We evaluate anomaly detection using AUC without optimizing it during classifier training.

\subsection{Quantization and model deployment}
\label{subsec:deployment}

We implement the models in TensorFlow and export them to TensorFlow Lite. NXP's Neutron converter~\cite{nxpNeutronConverter} prepares supported subgraphs for the eIQ Neutron NPU and exports model data as C arrays for deployment with TensorFlow Lite Micro~\cite{david2021tensorflow}. Quantization reduces the precision of supported operations to int8. Weights are quantized per output channel, while activations and quantized model interfaces use per-tensor scales. Operators and interfaces retained in float32 are excluded from this integer representation.

Deep SVDD uses quantization-aware training (QAT), with int8 input and float32 embedding output for distance scoring outside the graph. Its hypersphere center is fixed from the initial model and retained for both the float32 and quantized representations. AE, VAE, MAD, NF, and DDPM models use post-training quantization (PTQ). QAT conversion was limited by decoder and custom-layer support in our export pipeline. MC-Dropout uses PTQ in both experiments. DE use PTQ in Experiment~1 and QAT in Experiment~2. Where PTQ retained performance close to the float32 model, we used it without an additional QAT stage.

NPU execution depends on the exported graph. For MC-Dropout, the exported classifier combines an int8 convolutional backbone with a float32 head. The head runs on the CPU because our graph extractor did not support global pooling, concatenation, or unfolded batch normalization. Some exported graphs retain additional float32 operations. Consequently, an NPU result includes both NPU execution and CPU operations within the deployed graph. Deep SVDD distance calculations, reconstruction-error calculations, MAD aggregation, and negation of NF log-likelihoods are outside these graphs and are excluded from the hardware measurements.

The exported DDPM graphs implement a single denoising step. All tested DDPM configurations exceeded the SRAM limit of the MCXN947 evaluation kit, so DDPM is excluded from the reported results.

Predictive performance is evaluated on the PC for each model configuration and precision. These results are paired with board measurements of the corresponding deployed graphs. Predictive equivalence between PC and board execution is assumed and is not evaluated on the board.

\subsection{Energy and latency measurement protocol}
\label{subsec:measurement}

We measure execution on the MCXN947 evaluation kit~\cite{nxp2025mcxn947}, using an Arm Cortex-M33 core at 150\,MHz and the integrated eIQ Neutron NPU~\cite{bamberg2025eiq}. An MCU-Link Pro~\cite{nxp2023mculink} measures current on the 1.2\,V CORE supply domain, which powers the CPU and NPU. Reported energy is the incremental energy above an active-idle baseline on this domain. It does not represent the energy of the entire wearable system.

Per-score neural network energy and latency are estimated from the per-pass values using multipliers of $K$ for MC-Dropout, $M$ for DE, and 512 for MAD. The remaining methods use one pass. These estimates exclude scoring outside the deployed graphs, including UAD scoring, UQ prediction aggregation, and mutual-information calculation.

Each model is measured in five independent runs. Each run comprises 25 consecutive forward passes, with five repetitions for each of five input samples. The SysTick timer is used to measure the latency of each pass on-chip, and the durations are reported over UART independently of the current trace. For energy estimation, we identify the boundaries of the complete GPIO-gated burst in the current trace and partition it into per-pass sub-windows with lengths proportional to the UART-reported durations. Approximately one second of active-idle operation before and after the burst provides two baseline plateaus. The NPU clock state matches the execution mode. The run's baseline is the average of the two plateau median currents. For each pass, we subtract this baseline from its sub-window mean current and multiply by 1.2\,V and the independently measured duration to estimate its energy. We average the 25 per-pass values for each metric and report the median of the five run means.

Energy dispersion is the median of the five within-run standard deviations of per-pass energy estimates. For MC-Dropout, DE, and MAD, we scale this dispersion by $K$, $M$, and 512, respectively. This propagation assumes fully correlated measurement deviations and preserves relative energy dispersion. The underlying standard deviations describe observed variation across inputs and repetitions. They do not separately quantify systematic uncertainty from calibration, supply voltage, or baseline estimation. Latency dispersion is reported per forward pass without scaling. The scope of the code release is specified in the Code availability statement.

\section{Experiment 1: anomaly detection and execution cost}
\label{sect:experiment1}

The first experiment examines how well the methods distinguish admissible ECGs from anomalies and how much computation this distinction requires. We retain the four PTB-XL unseen-disease scenarios and the separate BUT QDB noise task from our previous work~\cite{ibrahim2026enhancing}. Deep SVDD, AE, VAE, MAD, and NF are evaluated in all five scenarios. MC-Dropout and DE are compared with these filters on the four disease scenarios. The standalone noise task has no underlying diagnostic classification task for these UQ baselines, as explained in Section~\ref{subsec:datasets}. We measure detection performance on the PC using the area under the receiver operating characteristic curve (AUC), treating the withheld disease or excessive noise as the positive class. This evaluates the ordering of anomaly scores without choosing a rejection threshold.

We pair these AUCs with the estimated neural network energy and latency per score obtained from board measurements, using the protocol in Section~\ref{subsec:measurement}. We also report SRAM usage, defined here as the memory used within each deployed graph's tensor arena. Other application memory is excluded. Values are given in decimal KB (1\,KB = 1000 bytes). The CPU comparison includes float32 and int8 deployments, while the NPU comparison uses int8 deployments. For each method and scenario, we exclude a configuration from the figures if an evaluated alternative of the same method has lower cost and equal or higher mean AUC. The energy and latency comparisons can therefore retain different configurations. AUC means and standard deviations are calculated across five training runs with fixed datasets. The error bars describe variability due to training seeds. Early stopping uses validation data, while the reported AUC maxima and budget comparisons summarize test-set performance. CPU execution provides a reference without differences in NPU operator coverage. We evaluate energy budgets separately for CPU and NPU deployments to examine the performance available under each execution setting.

\subsection{Results}

Table~\ref{tab:exp1_max} reports the highest observed mean AUC across the evaluated float32 and int8 configurations for each method and scenario. The accompanying SRAM values refer to CPU execution of the selected configuration. No method leads across all tasks. NF and MAD reach similar maxima for HYP, Deep SVDD leads for CD, and MC-Dropout leads for STTC and MI. All five UAD methods reach a noise-detection AUC of approximately 0.97 or higher. Among these selected configurations, SRAM usage ranges from 10.8\,KB for AE in noise detection to 256.4\,KB for AE in CD and MI. These maxima do not establish which configuration should be deployed, because their energy, latency, and memory requirements differ.

\begin{table}[pos=t]
  \centering
  \caption{Highest observed mean AUC and corresponding SRAM usage. Each maximum is selected separately per method and scenario across the evaluated float32 and int8 configurations, without an energy or latency constraint. AUC values show mean $\pm$ SD across five training seeds. Values in parentheses below the AUCs give SRAM usage in KB for CPU execution of the same configuration. UQ entries use the matching classifier reference graph. Bold identifies the highest displayed AUC per scenario without implying statistical significance. HYP, CD, STTC, and MI identify the withheld disease class. Noise denotes the BUT QDB task. A dash indicates that the UQ baseline is not applicable.}
  \label{tab:exp1_max}
  % Generated by analysis/build_experiment_tables.py; do not edit values.
\begin{tabular}{lccccc}
\toprule
Method & HYP & CD & STTC & MI & Noise \\
\midrule
Deep SVDD & 0.7792\,$\pm$\,0.0071 & \textbf{0.7658}\,$\pm$\,0.0065 & 0.6647\,$\pm$\,0.0346 & 0.6725\,$\pm$\,0.0044 & 0.9896\,$\pm$\,0.0022 \\
 & {\footnotesize (132.9\,KB)} & {\footnotesize (247.8\,KB)} & {\footnotesize (132.9\,KB)} & {\footnotesize (132.9\,KB)} & {\footnotesize (67.1\,KB)} \\
AE & 0.7812\,$\pm$\,0.0013 & 0.6547\,$\pm$\,0.0009 & 0.6145\,$\pm$\,0.0018 & 0.5903\,$\pm$\,0.0018 & \textbf{0.9903}\,$\pm$\,0.0012 \\
 & {\footnotesize (67.4\,KB)} & {\footnotesize (256.4\,KB)} & {\footnotesize (106.9\,KB)} & {\footnotesize (256.4\,KB)} & {\footnotesize (10.8\,KB)} \\
VAE & 0.7798\,$\pm$\,0.0010 & 0.6546\,$\pm$\,0.0005 & 0.6086\,$\pm$\,0.0028 & 0.5821\,$\pm$\,0.0021 & 0.9876\,$\pm$\,0.0017 \\
 & {\footnotesize (50.9\,KB)} & {\footnotesize (112.1\,KB)} & {\footnotesize (73.6\,KB)} & {\footnotesize (83.3\,KB)} & {\footnotesize (116.8\,KB)} \\
MAD & 0.8108\,$\pm$\,0.0008 & 0.7569\,$\pm$\,0.0047 & 0.6645\,$\pm$\,0.0054 & 0.6973\,$\pm$\,0.0061 & 0.9699\,$\pm$\,0.0012 \\
 & {\footnotesize (105.3\,KB)} & {\footnotesize (145.1\,KB)} & {\footnotesize (145.1\,KB)} & {\footnotesize (145.1\,KB)} & {\footnotesize (60.4\,KB)} \\
NF & \textbf{0.8113}\,$\pm$\,0.0004 & 0.7235\,$\pm$\,0.0022 & 0.6655\,$\pm$\,0.0029 & 0.6473\,$\pm$\,0.0024 & 0.9756\,$\pm$\,0.0009 \\
 & {\footnotesize (66.6\,KB)} & {\footnotesize (115.3\,KB)} & {\footnotesize (115.1\,KB)} & {\footnotesize (115.3\,KB)} & {\footnotesize (93.8\,KB)} \\
MC-Dropout & 0.6937\,$\pm$\,0.0067 & 0.6828\,$\pm$\,0.0051 & \textbf{0.7396}\,$\pm$\,0.0133 & \textbf{0.7064}\,$\pm$\,0.0056 & -- \\
 & {\footnotesize (198.9\,KB)} & {\footnotesize (198.9\,KB)} & {\footnotesize (198.9\,KB)} & {\footnotesize (198.9\,KB)} &  \\
DE & 0.6910\,$\pm$\,0.0037 & 0.6849\,$\pm$\,0.0070 & 0.7320\,$\pm$\,0.0062 & 0.6921\,$\pm$\,0.0055 & -- \\
 & {\footnotesize (198.9\,KB)} & {\footnotesize (198.9\,KB)} & {\footnotesize (209.9\,KB)} & {\footnotesize (198.9\,KB)} &  \\
\bottomrule
\end{tabular}

\end{table}

\begin{figure}[pos=t]
  \centering
  \includegraphics[width=\linewidth]{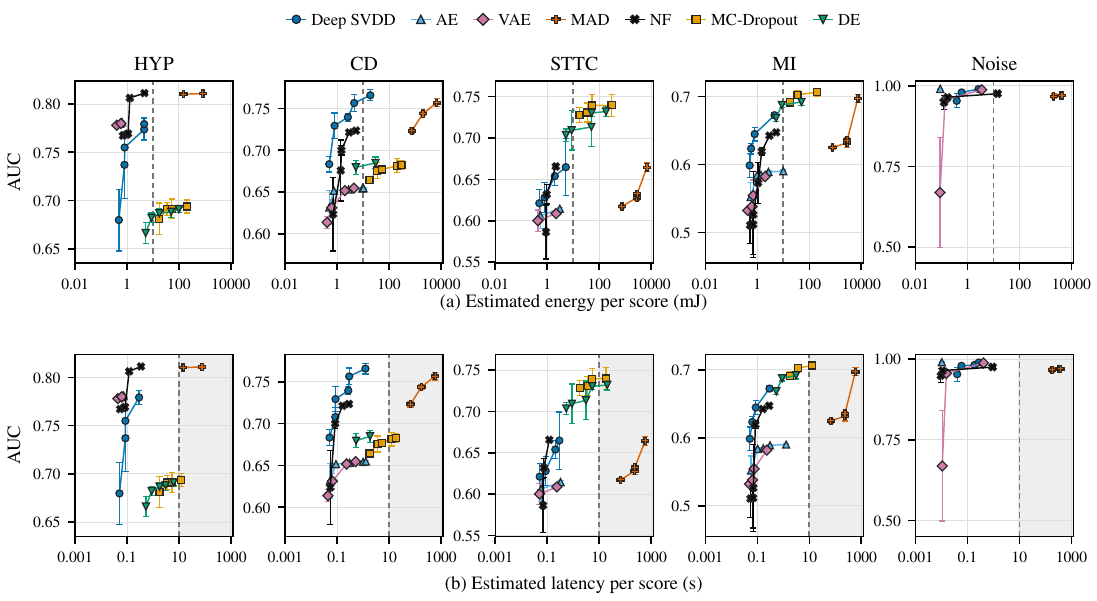}
  \caption{Experiment~1: mean detection AUC evaluated on the PC against estimated neural network energy (top) and latency (bottom) per score for CPU execution. Points show configurations retained separately for each method and cost metric. Error bars indicate one standard deviation across five training seeds. Cost axes are logarithmic. The energy reference at 10\,mJ represents an illustrative 1\,mW allocation for one score every 10\,s. Configurations in the shaded region beyond the 10\,s latency line cannot sustain sequential processing of one window every 10\,s.}
  \label{fig:exp1_cpu}
\end{figure}

\begin{figure}[pos=t]
  \centering
  \includegraphics[width=\linewidth]{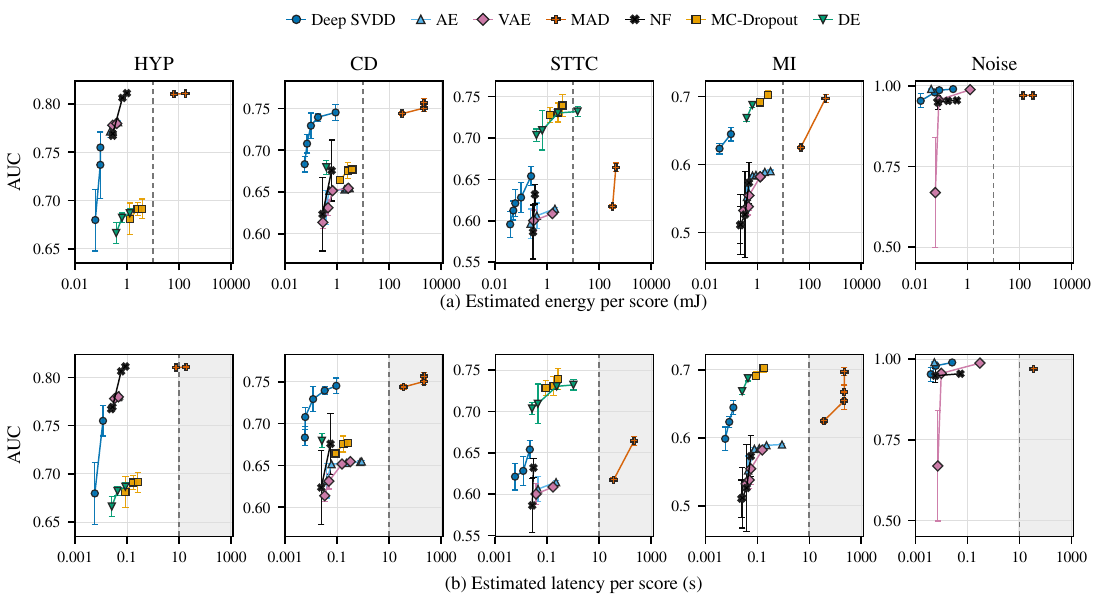}
  \caption{Experiment~1: mean detection AUC evaluated on the PC against estimated neural network energy (top) and latency (bottom) per score for NPU execution. Costs include CPU operations retained within the deployed graphs. Error bars show one standard deviation across five training seeds. Point selection, logarithmic cost axes, and the 10\,mJ and 10\,s references follow Fig.~\ref{fig:exp1_cpu}. The shaded region again identifies configurations too slow for sequential processing of one window every 10\,s.}
  \label{fig:exp1_npu}
\end{figure}

The lower rows of Figs.~\ref{fig:exp1_cpu} and~\ref{fig:exp1_npu} assess whether score computation can sustain sequential processing of one window every 10\,s. On the CPU, all plotted Deep SVDD, AE, VAE, and NF configurations finish within this interval. NF reaches its HYP maximum in 0.34\,s, while AE reaches its noise-detection maximum in approximately 10\,ms. The CPU UQ results span both sides of the reference line. In STTC, an int8 MC-Dropout configuration reaches an AUC of 0.739 in 5.36\,s, whereas the DE configuration with the highest mean AUC requires 20.48\,s. On the NPU, every plotted UQ configuration meets the interval. The same STTC DE configuration now requires 1.01\,s. These gains depend on each graph's operator coverage and retained CPU operations. Every plotted CPU MAD configuration exceeds the limit because each score requires 512 masked reconstructions. NPU execution brings a smaller MAD configuration below it in HYP, but not in the other scenarios.

Energy has no equivalent universal limit because a sensor patch must divide its available power among acquisition, processing, storage, and communication. Published examples include a wireless ECG patch consuming 1\,mW~\cite{shin2013ultra} and a multimodal patch consuming 7.4\,mW while recording ECG, seismocardiography, and environmental measurements to local storage~\cite{etemadi2016wearable}. These are device-level measurements, not inference allowances. The 10\,mJ lines in the upper rows of Figs.~\ref{fig:exp1_cpu} and~\ref{fig:exp1_npu} illustrate a 1\,mW scoring allocation for one score every 10\,s. This allocation equals the first patch's entire power consumption, so falling below the line does not establish complete-system feasibility. On the CPU, all plotted VAE configurations and most Deep SVDD, AE, and NF configurations fall below this reference. MAD and MC-Dropout remain above it, while DE spans both sides. At their respective CPU HYP maxima, NF requires less than 5\,mJ per score and MAD more than 800\,mJ. In CD, Deep SVDD approaches its CPU maximum below 5\,mJ, although reaching the maximum itself requires approximately 19\,mJ. AE reaches the highest noise-detection AUC with less than 0.1\,mJ per score. Within the CPU allowance of 10\,mJ, DE has the highest mean AUC in both STTC and MI. NPU acceleration brings all plotted MC-Dropout configurations below this allowance. For the int8 STTC MC-Dropout configuration described above, energy falls from 52.53\,mJ on the CPU to 3.85\,mJ on the NPU with unchanged AUC, bringing a higher-AUC alternative below the line. The lighter filters also benefit: all plotted NPU Deep SVDD and NF configurations require less than 1\,mJ per score. MAD remains above 10\,mJ on both backends.

Tables~\ref{tab:exp1_budget} and~\ref{tab:exp1_budget_npu} compare the detection performance available under CPU and NPU energy budgets, respectively. For each backend and budget, we select the highest observed mean AUC that satisfies the limit separately in each disease scenario, then average the four selected AUCs. On the CPU, Deep SVDD has the highest mean under all three finite budgets. At 10\,mJ, its mean AUC is 0.718, compared with 0.712 for NF and 0.690 for DE. DE can therefore cover all four scenarios within this budget, whereas MC-Dropout has no evaluated configuration below it. Increasing the allowance to 100\,mJ raises the DE mean to 0.699 and admits MC-Dropout at 0.703, both below Deep SVDD's 0.721. MAD has the highest unconstrained mean, but cannot cover all four scenarios within any of the three finite budgets. The unconstrained maxima therefore need to be considered alongside the energy required to achieve them.

\begin{table}[pos=t]
  \centering
  \caption{Best mean AUC $\pm$ propagated SD across the four PTB-XL disease scenarios under a per-score energy limit for CPU execution. Selection includes float32 and int8 deployments. Each scenario contributes its highest feasible mean AUC with equal weight and must separately satisfy the budget. The propagated SD describes training variability for fixed configurations, not the standard error of the mean. Bold identifies the highest displayed mean per column, without implying statistical significance. A dash indicates an absent feasible configuration in at least one scenario.}
  \label{tab:exp1_budget}
  % Generated by analysis/build_experiment_tables.py; do not edit values.
\begin{tabular}{lcccc}
\toprule
Method & 1 mJ & 10 mJ & 100 mJ & Unconstrained \\
\midrule
Deep SVDD & \textbf{0.6893}\,$\pm$\,0.0075 & \textbf{0.7183}\,$\pm$\,0.0092 & \textbf{0.7206}\,$\pm$\,0.0090 & 0.7206\,$\pm$\,0.0090 \\
AE & 0.6557\,$\pm$\,0.0039 & 0.6600\,$\pm$\,0.0008 & 0.6602\,$\pm$\,0.0008 & 0.6602\,$\pm$\,0.0008 \\
VAE & 0.6414\,$\pm$\,0.0069 & 0.6563\,$\pm$\,0.0009 & 0.6563\,$\pm$\,0.0009 & 0.6563\,$\pm$\,0.0009 \\
MAD & -- & -- & -- & \textbf{0.7324}\,$\pm$\,0.0023 \\
NF & 0.6375\,$\pm$\,0.0196 & 0.7119\,$\pm$\,0.0011 & 0.7119\,$\pm$\,0.0011 & 0.7119\,$\pm$\,0.0011 \\
MC-Dropout & -- & -- & 0.7025\,$\pm$\,0.0045 & 0.7056\,$\pm$\,0.0042 \\
DE & -- & 0.6896\,$\pm$\,0.0065 & 0.6988\,$\pm$\,0.0028 & 0.7000\,$\pm$\,0.0029 \\
\bottomrule
\end{tabular}

\end{table}

Table~\ref{tab:exp1_budget_npu} shows how the comparison changes for int8 NPU deployments. Deep SVDD leads at 1\,mJ, a budget that also admits DE in all four scenarios. At 10 and 100\,mJ, MC-Dropout has the highest mean AUC, at 0.703 compared with 0.700 for Deep SVDD. This small numerical difference does not establish statistical superiority. Deep SVDD already reaches its unconstrained int8 mean within 1\,mJ, so larger allowances do not improve its mean. MAD again has the highest unconstrained mean but cannot satisfy any tested finite budget in all four scenarios. Because CPU selection also permits float32, these comparisons reflect both execution cost and the available deployment precisions.

\begin{table}[pos=t]
  \centering
  \caption{Best mean AUC $\pm$ propagated SD across the four PTB-XL disease scenarios under a per-score energy limit for NPU execution. Selection uses int8 deployments. Each scenario contributes its highest feasible mean AUC with equal weight and must separately satisfy the budget. The propagated SD describes training variability for fixed configurations, not the standard error of the mean. Bold identifies the highest displayed mean per column, without implying statistical significance. A dash indicates an absent feasible configuration in at least one scenario.}
  \label{tab:exp1_budget_npu}
  % Generated by analysis/build_experiment_tables.py; do not edit values.
\begin{tabular}{lcccc}
\toprule
Method & 1 mJ & 10 mJ & 100 mJ & Unconstrained \\
\midrule
Deep SVDD & \textbf{0.6997}\,$\pm$\,0.0059 & 0.6997\,$\pm$\,0.0059 & 0.6997\,$\pm$\,0.0059 & 0.6997\,$\pm$\,0.0059 \\
AE & 0.6559\,$\pm$\,0.0039 & 0.6602\,$\pm$\,0.0008 & 0.6602\,$\pm$\,0.0008 & 0.6602\,$\pm$\,0.0008 \\
VAE & 0.6465\,$\pm$\,0.0065 & 0.6563\,$\pm$\,0.0009 & 0.6563\,$\pm$\,0.0009 & 0.6563\,$\pm$\,0.0009 \\
MAD & -- & -- & -- & \textbf{0.7324}\,$\pm$\,0.0023 \\
NF & 0.6730\,$\pm$\,0.0122 & 0.6730\,$\pm$\,0.0122 & 0.6730\,$\pm$\,0.0122 & 0.6730\,$\pm$\,0.0122 \\
MC-Dropout & -- & \textbf{0.7025}\,$\pm$\,0.0045 & \textbf{0.7025}\,$\pm$\,0.0045 & 0.7025\,$\pm$\,0.0045 \\
DE & 0.6896\,$\pm$\,0.0065 & 0.6960\,$\pm$\,0.0028 & 0.6964\,$\pm$\,0.0030 & 0.6964\,$\pm$\,0.0030 \\
\bottomrule
\end{tabular}

\end{table}

SRAM usage provides an additional distinction between the methods. At their HYP maxima, NF and MAD use 66.6 and 105.3\,KB of SRAM for CPU execution, despite their approximately 179-fold energy difference. NPU execution can also increase SRAM usage. For the same int8 Deep SVDD CD graph, moving from CPU to NPU execution reduces energy from 2.60 to 0.18\,mJ while increasing SRAM usage from 73.3 to 84.8\,KB.

\section{Experiment 2: rejection and classification}
\label{sect:experiment2}

The second experiment examines whether anomaly-based rejection improves the decisions of the complete monitoring pipeline and how this benefit relates to execution energy. We use the four PTB-XL unseen-disease scenarios with added electrode-motion noise, as described in Section~\ref{subsec:datasets}. An input is considered invalid if it contains an unseen disease, added noise, or both. We compare a two-stage system comprising a Deep SVDD filter followed by a diagnostic classifier with MC-Dropout and DE. This system follows the upstream path in Fig.~\ref{fig:system}, with energy $E_{2\mathrm{s}}$ in Eq.~\eqref{eq:filter_energy}. Deep SVDD is retained for its favorable balance between detection performance and execution cost in Experiment~1, including the lowest tested energy budget on both backends. Classifier widths of $w=0.25$ and $w=1.0$ allow us to examine how the comparison changes when classification becomes more expensive. The narrower network provides the lower-cost setting for the UQ baselines.

We use the custom accuracy of our previous work~\cite{ibrahim2026enhancing} to evaluate both rejection and classification. A decision is correct when an admissible input is accepted and all four retained binary diagnostic labels are predicted correctly, or when an invalid input is rejected. All other outcomes are incorrect. Custom accuracy $A$ is the fraction of correct decisions among all input windows, including rejected windows. It therefore differs from classification accuracy among retained recordings. Diagnostic decision thresholds are fitted on the clean ID validation subset. Each label's threshold is initialized at its F1-optimal point on the precision--recall curve, then refined through up to five passes of coordinate ascent on four-label exact-match accuracy. These thresholds are applied unchanged to the test data.

We vary the anomaly-score rejection threshold to examine the trade-off between withholding unsuitable inputs and discarding useful recordings. Custom accuracy is evaluated on the PC. Curves report its mean and one standard deviation across five training seeds, with the dataset and noise realizations fixed. Energy per incoming window is estimated from board measurements using Eqs.~\eqref{eq:filter_energy} and~\eqref{eq:uq_energy}. We also report $E/A$, the estimated energy per correct system decision. Correct rejections contribute to this denominator, so this quantity is not energy per correct diagnosis.

The energy tables report propagated hardware-measurement dispersion, with accuracy variability also included for $E/A$. Appendix~\ref{app:dispersion} describes this propagation.

Early stopping and diagnostic label thresholds use validation data exclusively. Model configurations are selected using the test results from Experiment~1 and then evaluated here using custom accuracy. The highlighted accuracy maxima are oracle operating points selected retrospectively from the test-set rejection sweep. A deployed system could not identify them without labeled data. We apply this rule consistently to all three methods, but the absolute performance estimates remain optimistic and do not constitute an independent evaluation of a prospectively selected rejection threshold. Comparisons use each method's own accuracy-maximizing rejection rate. Appendix~\ref{app:configurations} lists the selected configurations.

\subsection{Results}

\begin{figure}[pos=t]
  \centering
  \includegraphics[width=\linewidth]{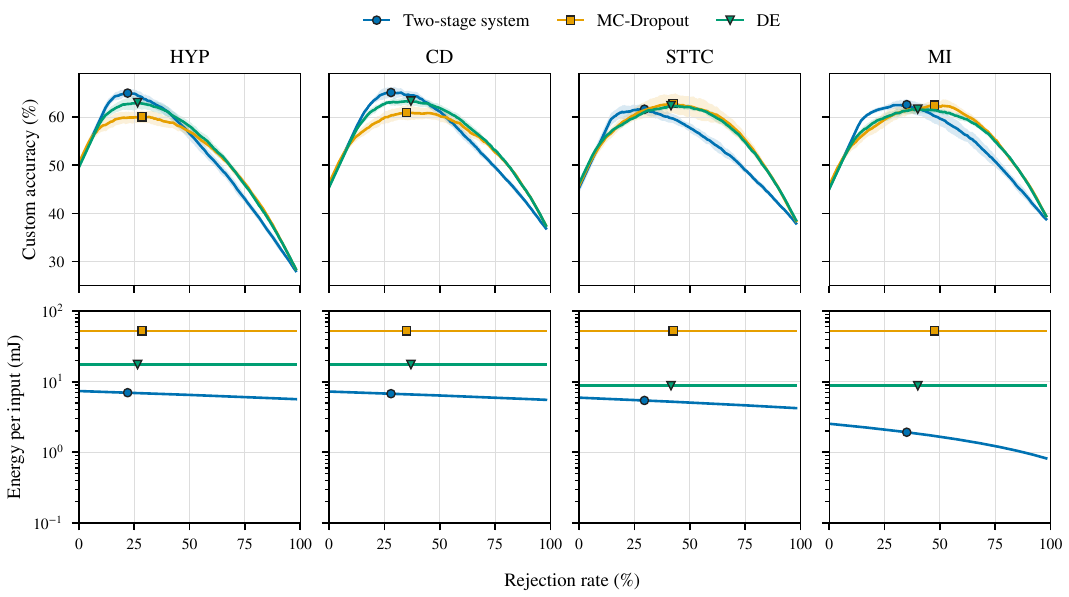}
  \caption{Experiment~2 with CPU execution costs and classifier width $w=0.25$. The upper row shows custom accuracy evaluated on the PC, and the lower row shows estimated neural network energy per incoming window on a logarithmic scale. Columns identify the withheld disease class. Shading indicates one standard deviation across five training seeds. In both rows, markers identify each method's maximum mean custom accuracy. Their energy values therefore refer to the same operating points as their accuracy values.}
  \label{fig:exp2_cpu025}
\end{figure}

Figure~\ref{fig:exp2_cpu025} shows why detection performance must be evaluated together with downstream decisions. Initially, rejection increases custom accuracy by withholding inputs that the classifier should not interpret. Beyond the accuracy peak, further rejection increasingly removes admissible recordings that could have been classified correctly. The two-stage system raises mean custom accuracy by 15.0--19.4 percentage points relative to the same classifier without rejection across the four scenarios. Its peak accuracy exceeds both UQ baselines in HYP and CD, while the peak values are closer across methods in STTC and MI. Its maxima also occur at lower rejection rates than those of either UQ baseline in all four scenarios. In the lower row, the two-stage system remains less costly than both UQ methods throughout the sweep, with a larger gap to MC-Dropout than to DE. This gap increases with rejection because fewer windows reach the second-stage classifier. UQ energy remains constant because all classifier evaluations are required before rejection can occur.

Table~\ref{tab:exp2_peaks} reports each method's results at the rejection rate that maximizes its mean custom accuracy in Fig.~\ref{fig:exp2_cpu025}. The two-stage system reaches the highest numerical peak in HYP, CD, and MI. The MI result is close to MC-Dropout, at $62.58\pm1.06\%$ and $62.49\pm0.92\%$, respectively, where the spreads are SDs across training seeds. MC-Dropout has the highest peak in STTC. At their respective accuracy peaks, the two-stage system requires 1.93--6.99\,mJ per input, compared with 8.76--17.51\,mJ for DE and 52.53\,mJ for MC-Dropout. Its energy per correct decision is also lower in all four scenarios. These comparisons do not match the methods at identical accuracy or rejection rates.

\begin{table}[pos=t]
  \centering
  \caption{Accuracy evaluated on the PC and estimated CPU energy costs at $w=0.25$ and each method's oracle accuracy peak, selected from the test-set rejection sweep. $A^*$ is mean custom accuracy $\pm$ SD across five training seeds, and $r^*$ is the rejected fraction. $E^*$ (energy per input) and $E^*/A^*$ (energy per correct decision) show estimates $\pm$ propagated dispersion. The latter includes accuracy variability. Bold identifies the highest accuracy and lowest costs per scenario, without implying statistical significance.}
  \label{tab:exp2_peaks}
  % Generated by analysis/build_experiment_tables.py; do not edit values.
\begin{tabular}{llrrrr}
\toprule
Scenario & Method & $A^*$ (\%) & $r^*$ (\%) & $E^*$ (mJ/input) & $E^*/A^*$ (mJ/correct) \\
\midrule
HYP & Two-stage system & \textbf{64.98}\,$\pm$\,0.86 & 22.0 & \textbf{6.99}\,$\pm$\,0.06 & \textbf{10.75}\,$\pm$\,0.17 \\
 & MC-Dropout & 60.06\,$\pm$\,1.20 & 28.5 & 52.53\,$\pm$\,0.96 & 87.46\,$\pm$\,2.37 \\
 & DE & 62.98\,$\pm$\,1.28 & 26.5 & 17.51\,$\pm$\,0.32 & 27.80\,$\pm$\,0.76 \\
\midrule
CD & Two-stage system & \textbf{65.10}\,$\pm$\,1.09 & 28.0 & \textbf{6.76}\,$\pm$\,0.05 & \textbf{10.39}\,$\pm$\,0.19 \\
 & MC-Dropout & 60.92\,$\pm$\,1.54 & 35.0 & 52.53\,$\pm$\,0.96 & 86.23\,$\pm$\,2.69 \\
 & DE & 63.34\,$\pm$\,0.79 & 37.0 & 17.51\,$\pm$\,0.32 & 27.65\,$\pm$\,0.61 \\
\midrule
STTC & Two-stage system & 61.63\,$\pm$\,1.10 & 29.5 & \textbf{5.43}\,$\pm$\,0.04 & \textbf{8.80}\,$\pm$\,0.17 \\
 & MC-Dropout & \textbf{62.81}\,$\pm$\,1.95 & 42.5 & 52.53\,$\pm$\,0.96 & 83.64\,$\pm$\,3.01 \\
 & DE & 62.38\,$\pm$\,0.58 & 41.5 & 8.76\,$\pm$\,0.16 & 14.04\,$\pm$\,0.29 \\
\midrule
MI & Two-stage system & \textbf{62.58}\,$\pm$\,1.06 & 35.0 & \textbf{1.93}\,$\pm$\,0.04 & \textbf{3.08}\,$\pm$\,0.08 \\
 & MC-Dropout & 62.49\,$\pm$\,0.92 & 47.5 & 52.53\,$\pm$\,0.96 & 84.06\,$\pm$\,1.97 \\
 & DE & 61.63\,$\pm$\,0.83 & 40.0 & 8.76\,$\pm$\,0.16 & 14.21\,$\pm$\,0.32 \\
\bottomrule
\end{tabular}

\end{table}

\begin{figure}[pos=t]
  \centering
  \includegraphics[width=\linewidth]{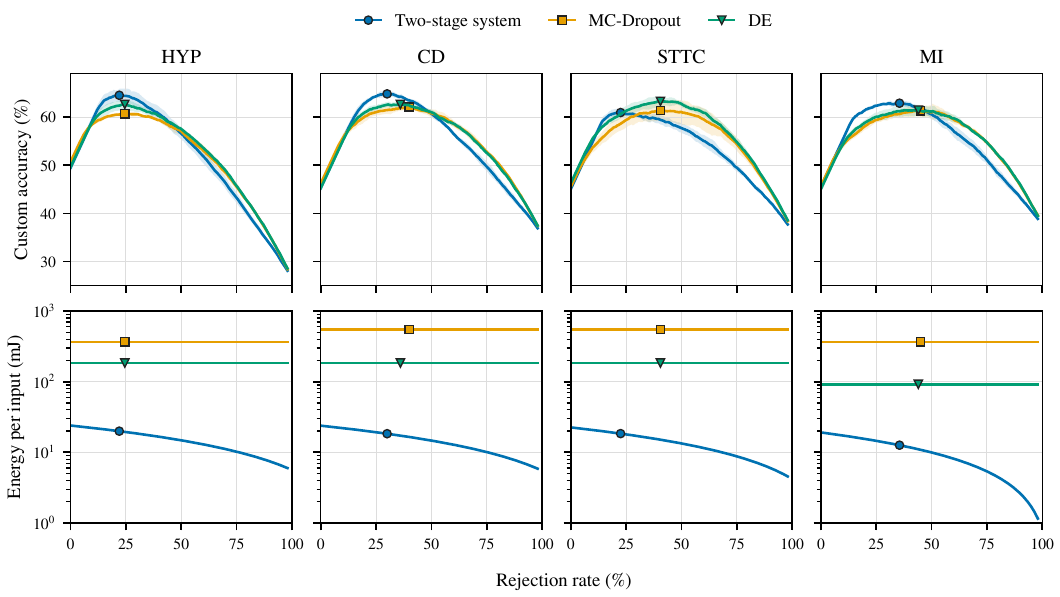}
  \caption{Experiment~2 with CPU execution costs and classifier width $w=1.0$, using the same filter configurations as in Fig.~\ref{fig:exp2_cpu025}. Accuracy, shading, and peak markers follow the same conventions. The wider classifier increases both accepted-window classification cost and the cost of repeated UQ evaluations. All UQ configurations shown exceed 10\,s of graph execution per input on the CPU.}
  \label{fig:exp2_cpu100}
\end{figure}

Figure~\ref{fig:exp2_cpu100} examines the effect of increasing classifier width to $w=1.0$. Peak custom accuracy does not improve consistently. The two-stage system retains the highest numerical peaks in HYP, CD, and MI, while DE now has the highest peak in STTC. The energy differences are larger for the selected full-width configurations. At the respective accuracy peaks, DE requires approximately 7.3--10 times the energy per input of the two-stage system, while MC-Dropout requires 18--30 times as much. The wider classifier also makes the two-stage energy curve decrease more steeply with rejection, because each rejected window avoids a more costly classifier evaluation. The filter cost remains unchanged, so the filter-to-classifier energy ratio $\rho$ decreases. This strengthens the relative energy advantage of the two-stage system, as predicted by Eq.~\eqref{eq:energy_break_even}. The full-width CPU UQ configurations also exceed the 10\,s input cadence: MC-Dropout requires approximately 41--61\,s per input and DE 10.24--20.48\,s. The two-stage system requires at most 2.66\,s of graph execution for an accepted input. These estimates exclude computation outside the deployed graphs.

Increasing classifier width also increases SRAM usage. The classifier reference graph requires 54.4\,KB at $w=0.25$ and 209.9\,KB at $w=1.0$ for CPU execution, compared with 60.0 and 219.7\,KB for NPU execution. The selected Deep SVDD filters require 39.5--216.0\,KB for CPU execution and 43.9--270.2\,KB for NPU execution, with the same filter configurations used at both classifier widths. These values describe individual deployed graphs and do not establish the combined SRAM requirements of the complete pipelines.

Table~\ref{tab:exp2_daily} expresses the accuracy-peak energy estimates as a repeated monitoring workload: 8,640 ten-second windows, corresponding to one window every 10\,s for one day. The two-stage system has the lowest estimated energy in every listed scenario and hardware setting. NPU execution reduces the absolute costs, although the benefit differs between the filter and classifier graphs. The full-width CPU UQ entries describe only the arithmetic energy of this workload because their latency prevents completion at the assumed cadence. None of these values represents total daily device energy. Sensing, idle operation, communication, scoring outside the graphs, and escalation are excluded. The NPU accuracy--energy curves are provided in Appendix~\ref{app:npu}.

\begin{table}[pos=t]
  \centering
  \caption{Extrapolated neural network energy in joules for 8,640 input windows at each method's own maximum mean custom accuracy. Both per-input energy and its propagated dispersion are multiplied by 8,640. The resulting $\pm$ values describe dispersion attached to the extrapolation, not measured day-to-day variability. The workload corresponds to ten-second windows arriving every 10\,s for one day. Bold identifies the lowest estimated energy within each hardware setting and scenario. Other wearable-system energy costs are excluded.}
  \label{tab:exp2_daily}
  % Generated by analysis/build_experiment_tables.py; do not edit values.
\begin{tabular}{llrrrr}
\toprule
Setting & Method & HYP (J) & CD (J) & STTC (J) & MI (J) \\
\midrule
CPU, $w=0.25$ & Two-stage system & \textbf{60.4}\,$\pm$\,0.5 & \textbf{58.4}\,$\pm$\,0.4 & \textbf{46.9}\,$\pm$\,0.4 & \textbf{16.6}\,$\pm$\,0.3 \\
 & MC-Dropout & 453.9\,$\pm$\,8.3 & 453.9\,$\pm$\,8.3 & 453.9\,$\pm$\,8.3 & 453.9\,$\pm$\,8.3 \\
 & DE & 151.3\,$\pm$\,2.8 & 151.3\,$\pm$\,2.8 & 75.6\,$\pm$\,1.4 & 75.6\,$\pm$\,1.4 \\
\midrule
CPU, $w=1$ & Two-stage system & \textbf{172.2}\,$\pm$\,0.8 & \textbf{158.5}\,$\pm$\,0.7 & \textbf{159.1}\,$\pm$\,0.7 & \textbf{109.1}\,$\pm$\,0.6 \\
 & MC-Dropout & 3171.1\,$\pm$\,16.3$^{\dagger}$ & 4756.7\,$\pm$\,24.5$^{\dagger}$ & 4756.7\,$\pm$\,24.5$^{\dagger}$ & 3171.1\,$\pm$\,16.3$^{\dagger}$ \\
 & DE & 1585.6\,$\pm$\,8.2$^{\dagger}$ & 1585.6\,$\pm$\,8.2$^{\dagger}$ & 1585.6\,$\pm$\,8.2$^{\dagger}$ & 792.8\,$\pm$\,4.1$^{\dagger}$ \\
\midrule
NPU, $w=0.25$ & Two-stage system & \textbf{8.3}\,$\pm$\,1.3 & \textbf{8.4}\,$\pm$\,1.3 & \textbf{3.2}\,$\pm$\,0.9 & \textbf{1.5}\,$\pm$\,0.3 \\
 & MC-Dropout & 33.3\,$\pm$\,10.5 & 33.3\,$\pm$\,10.5 & 33.3\,$\pm$\,10.5 & 33.3\,$\pm$\,10.5 \\
 & DE & 11.1\,$\pm$\,3.5 & 11.1\,$\pm$\,3.5 & 5.5\,$\pm$\,1.8 & 5.5\,$\pm$\,1.8 \\
\midrule
NPU, $w=1$ & Two-stage system & \textbf{17.4}\,$\pm$\,1.4 & \textbf{16.5}\,$\pm$\,1.4 & \textbf{12.3}\,$\pm$\,1.1 & \textbf{9.1}\,$\pm$\,0.6 \\
 & MC-Dropout & 255.3\,$\pm$\,17.4 & 382.9\,$\pm$\,26.0 & 382.9\,$\pm$\,26.0 & 255.3\,$\pm$\,17.4 \\
 & DE & 127.6\,$\pm$\,8.7 & 127.6\,$\pm$\,8.7 & 127.6\,$\pm$\,8.7 & 63.8\,$\pm$\,4.3 \\
\bottomrule
\end{tabular}

  \par\smallskip
  \begin{minipage}{\linewidth}
    \footnotesize\raggedright
    $^{\dagger}$ CPU UQ configurations whose graph latency exceeds 10\,s, preventing completion at the assumed cadence.
  \end{minipage}
\end{table}

\section{Discussion}
\label{sect:discussion}

This study examines whether a separate anomaly detector can improve ECG monitoring decisions at a lower computational cost than classifier-based UQ. We extend our previous work by comparing upstream UAD with MC-Dropout and DE, measuring deployed graph execution, and relating these costs to rejection and classification. The combined results allow us to distinguish two questions: which methods detect unsuitable inputs within a resource budget, and when their use improves system decisions and reduces estimated neural network energy.

\subsection{Detecting unseen diseases and noise}
\label{subsec:discussion_detection}

The disease-specific rankings show that detection difficulty depends on the method, not only on the withheld diagnosis. The highest observed AUCs are achieved by UAD methods in HYP and CD and by MC-Dropout in STTC and MI. In particular, STTC has the highest AUC among the four disease scenarios for both UQ methods, despite being relatively difficult for UAD. Our previous discussion attributed part of the difficulty in STTC and MI to overlapping ECG morphology~\cite{ibrahim2026enhancing}. This remains plausible. For example, ventricular hypertrophy can produce secondary ST--T abnormalities~\cite{hancock2009aha}, so withholding a diagnostic label need not remove every related morphological feature from training. However, we exclude all recordings carrying the withheld label, including recordings with additional diagnoses. We have not established which overlapping features remain in the retained data or whether they explain the observed rankings.

The distinction between input modeling and diagnostic prediction offers a complementary explanation. AE and VAE minimize reconstruction error, which need not increase for an unfamiliar disease if its waveform can still be reconstructed. Accurate reconstruction of anomalies has been demonstrated outside ECG analysis~\cite{bouman2025autoencoders}. Similarly, work on image data shows that NF likelihood can emphasize local correlations that do not distinguish familiar from unfamiliar semantic content~\cite{kirichenko2020why}. Deep SVDD instead evaluates distance in a feature space trained to concentrate admissible inputs near a fixed center. MC-Dropout and DE assess disagreement about the retained diagnostic labels, so they may respond to changes that matter for classification even when reconstruction or likelihood remains uninformative. Conversely, identical predicted probabilities across evaluations yield a zero mutual-information score, even if the input is unfamiliar. These mechanisms are possible explanations, not causes established by our experiments. The different rankings also agree with prior ECG evidence that OOD performance depends on both the uncertainty score and the disease class~\cite{barandas2024evaluation}.

Noise detection presents a different distinction. The BUT QDB task separates severe signal degradation from recordings in which QRS complexes remain reliably detectable~\cite{nemcova2020butqdb}. This may produce a more pronounced change in signal structure than an unseen diagnosis within an otherwise interpretable ECG. All evaluated UAD methods perform well on this task, and AE combines the highest observed AUC with a low execution cost. A reconstruction-based method can therefore remain useful even when it performs poorly in some disease scenarios. However, the noise and disease tasks differ in dataset, lead count, preprocessing, and anomaly definition. Their AUCs do not establish that noise is intrinsically easier to detect. The absence of an underlying classification task also prevents a comparison with the UQ baselines on BUT QDB.

Hardware measurements further qualify the method rankings. Similar detection performance can require very different execution costs, as illustrated by NF and MAD in HYP. MAD's repeated reconstructions impose a substantial energy and latency cost without a proportional increase in the SRAM required by one deployed graph. Parameter count is also an incomplete proxy. The Deep SVDD configuration selected for STTC in Experiment~2 has more parameters than the HYP configuration, yet requires less energy, latency, and SRAM on both backends. Runtime memory includes intermediate tensors and operator buffers, whose requirements depend on the graph and its implementation~\cite{david2021tensorflow}. Acceleration introduces another trade-off: the matched Deep SVDD CD example in Section~\ref{sect:experiment1} consumes less energy on the NPU while requiring more SRAM. These observations support measuring the relevant deployment costs separately, rather than inferring them from model size~\cite{dehghani2022efficiency}.

Within the evaluated configurations, Deep SVDD offers the strongest overall compromise for a CPU energy constraint. It has the highest mean AUC under every finite budget in Table~\ref{tab:exp1_budget}, favorable latency, and strong noise detection. This supports its selection for the two-stage system, but does not make it the preferred detector for every scenario. AE is attractive for the standalone noise task, and MC-Dropout provides higher disease-detection AUCs in STTC and MI. On the NPU, Deep SVDD leads under the tightest tested budget, while MC-Dropout has a slightly higher numerical mean under the two larger budgets (Table~\ref{tab:exp1_budget_npu}). Acceleration makes repeated classifier evaluations more affordable, but the available precision also matters: the NPU pool uses int8, whereas the CPU pool also includes float32. The preferred method therefore depends on both the energy allowance and the available deployments. These conclusions depend on the searched architectures, score definitions, and deployment toolchain. The failure to deploy the tested DDPM configurations establishes a limitation of those configurations on this board, not of diffusion-based detection in general.

\subsection{Implications for end-to-end monitoring}
\label{subsec:discussion_system}

Better anomaly-score ordering does not by itself determine the best monitoring system. Custom accuracy depends on which invalid inputs are rejected, which admissible inputs are retained, and whether the retained inputs receive the correct diagnostic labels. The two-stage system has higher numerical accuracy peaks in HYP and CD, while UQ is more competitive in STTC and MI. The comparison therefore reflects both rejection behavior and the predictions of the underlying classifiers. Experiment~2 supports Deep SVDD as an effective upstream filter in the tested settings, but evaluates only this UAD method downstream. It does not establish that the other UAD methods would preserve their Experiment~1 rankings after classification.

The two-stage system uses less estimated energy than either UQ method at their respective accuracy peaks in every evaluated hardware and width setting. This advantage follows from where rejection occurs. The two-stage system evaluates a relatively inexpensive filter on every input and invokes the classifier only after acceptance. MC-Dropout and DE complete their repeated classifier evaluations before making a rejection decision. For the selected configurations, increasing classifier width strengthens the relative advantage of the upstream filter, as expected from Eq.~\eqref{eq:energy_break_even}, without consistently improving accuracy. The selected full-width CPU UQ configurations also fail to sustain the input cadence, whereas the two-stage system remains fast enough. Thus, the classifier cost and available acceleration should enter the choice of rejection mechanism alongside detection performance. The comparisons at each method's own accuracy peak describe retrospective operating points observed in this experiment, rather than energy savings at identical accuracy or rejection rates.

An upstream filter does not necessarily reduce energy relative to classification without rejection. Against a single classifier pass, the energy saved by avoiding some classifications must exceed the cost of evaluating the filter on every input. This condition is not met at the reported accuracy peaks for the narrow CPU classifier. The additional expenditure buys a rejection mechanism, while the energy savings reported against MC-Dropout and DE arise from avoiding repeated classification. A consequential limitation is the absence of single-pass UQ alternatives. Evidential uncertainty and uncertainty-guided early exits have already been implemented on microcontrollers~\cite{jia2024ur2m,jia2026cascade}. Their inclusion could change the energy comparison. The present results therefore support an advantage over the evaluated repeated-inference methods, rather than over UQ generally.

Rejection also needs an operational meaning. A correctly rejected recording with an unseen disease remains clinically unresolved. A monitoring system would need a policy for reacquisition, transmission, or clinical review, and the corresponding costs could change the preferred operating point. Equation~\eqref{eq:escalation_energy} makes this dependence explicit. Custom accuracy gives equal credit to a correct rejection and a correct classification, although their clinical consequences differ. Its value depends on the prevalence and composition of the invalid inputs, and energy per correct decision inherits this dependence. These metrics assess the specified decision problem. They do not establish diagnostic benefit, an acceptable missed-event rate, or the workload created by escalation.

The reported operating points also require prospective validation. Configurations are selected from Experiment~1 test results, and accuracy peaks are selected from the Experiment~2 test sweep. Applying the same selection rule to all methods does not remove this optimism or ensure that relative rankings would persist on new data. The five-run spreads describe training-seed variability for fixed recordings and noise realizations. They do not measure uncertainty from sampling new patients, devices, or artifact conditions, and close numerical differences should not be interpreted as established superiority. A deployment comparison should select configurations and rejection thresholds on validation data, then evaluate them on an independent test cohort. Naturally corrupted wearable recordings would also test whether the conclusions extend beyond the controlled electrode-motion corruption and the combined noise-and-disease rejection task used here. The evaluation was not stratified by sex or gender. Generalizability across these groups remains to be assessed.

Finally, the measured costs describe deployed graph execution on one hardware platform. Repetition-based costs are extrapolated from a single measured pass, and scoring outside the graphs is excluded for every method. The SRAM values describe individual graphs, not the combined requirements of the monitoring pipelines. These measurements support the relative computational comparisons, but do not establish complete-system battery life or memory feasibility. Measuring the integrated pipeline, including escalation, and comparing single-pass alternatives would determine whether the observed benefit of separating rejection from classification persists under the full requirements of continuous monitoring.

\section{Conclusion}
\label{sect:conclusion}

We evaluated independent upstream UAD filters for rejecting unsuitable ECG inputs before diagnostic classification. The study combined comparisons with MC-Dropout and DE, measurements of graph execution energy, latency, and SRAM usage, and energy equations relating rejection to avoided classifier evaluations. Detection performance varied with the anomaly type, while Deep SVDD achieved the highest mean disease-detection AUC under each finite CPU energy budget tested. Among int8 NPU deployments, Deep SVDD led at the lowest tested budget, while MC-Dropout had slightly higher means at the larger budgets. The Deep SVDD two-stage system improved custom accuracy relative to classification without rejection. It also required less estimated neural network energy than both UQ baselines when each method was evaluated at its own retrospective accuracy peak. For the selected configurations, this relative energy advantage increased with classifier width. These findings support Deep SVDD as an efficient upstream filter within the evaluated settings, while showing that detector choice depends on the task and deployment constraints.

For remote patient monitoring, the main implication is that rejection should be designed jointly with classification and the available hardware resources. A lightweight filter can avoid expensive diagnostic evaluations, but its value depends on which recordings it withholds and how those recordings are handled. Future work should compare single-pass UQ alternatives and evaluate prospectively selected configurations and thresholds on independent, naturally corrupted wearable recordings. Measurements of the integrated pipeline should include sensing, scoring, and the cost of reacquisition or further assessment. This would establish whether the observed computational advantage persists when rejection becomes part of an operational monitoring workflow.

\appendix
\counterwithin{table}{section}
\counterwithin{figure}{section}
\section{Additional experimental details}
\subsection{Dispersion estimates}
\label{app:dispersion}

The energy dispersions in Tables~\ref{tab:exp2_peaks} and~\ref{tab:exp2_daily} originate from the hardware measurements described in Section~\ref{subsec:measurement}. For the two-stage system, the filter dispersion and $(1-r)$ times the classifier dispersion are combined in quadrature. First-order propagation for $E/A$ combines energy dispersion with the training-seed SD of accuracy. Both calculations hold the rejection rate fixed and assume zero covariance between the component terms. These dispersions describe the estimated average costs at that operating point. They do not include variability between individual acceptance and rejection events. For daily extrapolations, both the per-input energy estimate and its attached dispersion are scaled linearly by 8,640.

\subsection{Experiment 2 configurations}
\label{app:configurations}

Table~\ref{tab:exp2_configurations} identifies the configurations used in Experiment~2. The Deep SVDD architecture for each scenario is fixed across classifier widths and backends. Both processing paths use the int8 deployments described in Section~\ref{subsec:deployment}. Within each width and backend, classifier energy is represented by the same measured single-pass reference cost for all scenarios, consistent with the equal-cost approximation in Eq.~\eqref{eq:uq_energy}.

\begin{table}[pos=htbp]
  \centering
  \caption{Selected Experiment~2 configurations, shared between CPU and NPU deployments. The Parameters column gives the Deep SVDD parameter count. $K$ and $p$ are the number of MC-Dropout evaluations and dropout probability. $M$ is the number of ensemble members. Configurations were selected using Experiment~1 test results.}
  \label{tab:exp2_configurations}
  % Generated by analysis/build_experiment_tables.py; do not edit values.
\begin{tabular}{llrrrr}
\toprule
Width $w$ & Scenario & Parameters & $K$ & $p$ & $M$ \\
\midrule
0.25 & HYP & 89,344 & 30 & 0.5 & 10 \\
 & CD & 89,344 & 30 & 0.3 & 10 \\
 & STTC & 140,672 & 30 & 0.3 & 5 \\
 & MI & 15,360 & 30 & 0.5 & 5 \\
\midrule
1 & HYP & 89,344 & 20 & 0.3 & 10 \\
 & CD & 89,344 & 30 & 0.5 & 10 \\
 & STTC & 140,672 & 30 & 0.4 & 10 \\
 & MI & 15,360 & 20 & 0.3 & 5 \\
\bottomrule
\end{tabular}

\end{table}

\subsection{Experiment 2 NPU execution results}
\label{app:npu}

\begin{figure}[pos=t]
  \centering
  \includegraphics[width=\linewidth]{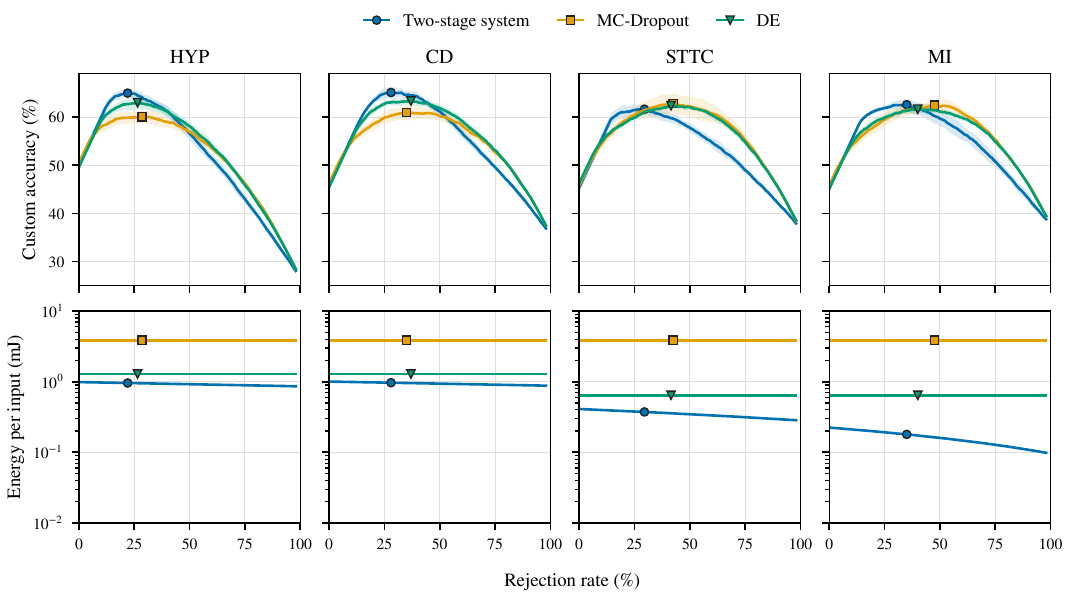}
  \caption{Experiment~2 with NPU execution costs and classifier width $w=0.25$. The upper row shows mean custom accuracy evaluated on the PC with one standard deviation across five training seeds. The lower row shows estimated neural network energy per incoming window, including CPU operations retained within the deployed graphs. Markers identify each method's accuracy peak in both rows.}
  \label{fig:exp2_npu025}
\end{figure}

\begin{figure}[pos=t]
  \centering
  \includegraphics[width=\linewidth]{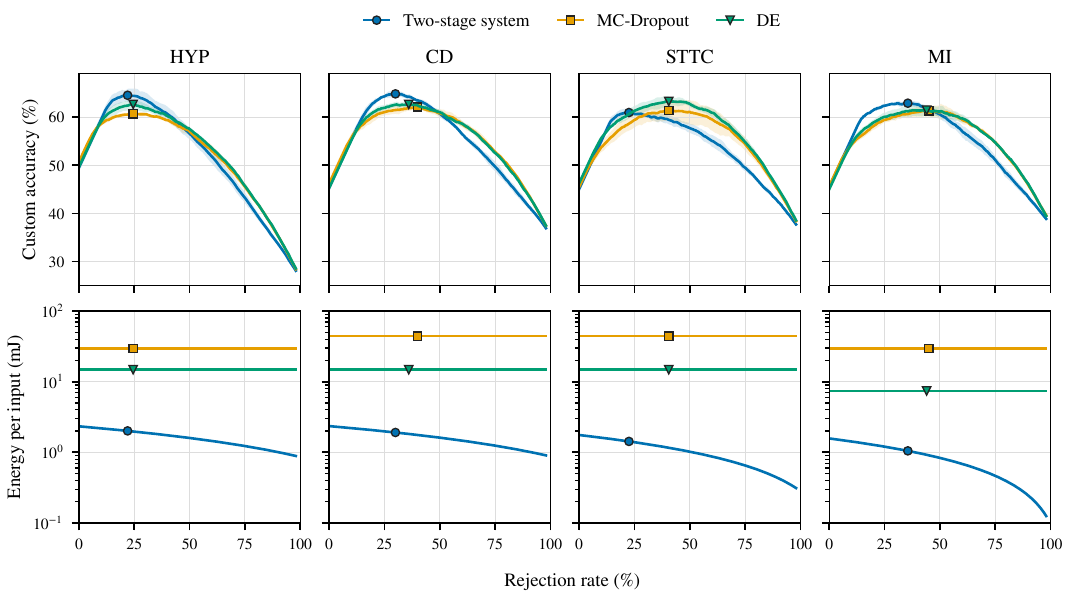}
  \caption{Experiment~2 with NPU execution costs and classifier width $w=1.0$. Plot conventions match Fig.~\ref{fig:exp2_npu025}. The filter configurations are unchanged, while the larger classifier increases the classification and UQ costs.}
  \label{fig:exp2_npu100}
\end{figure}

Figures~\ref{fig:exp2_npu025} and~\ref{fig:exp2_npu100} pair the same PC predictive results used in the CPU comparison with the corresponding NPU execution costs. The two-stage system retains an energy advantage at each method's accuracy peak, but the relative advantage is smaller for the narrow classifier on the NPU than on the CPU. All selected NPU configurations remain below 10\,s of estimated graph execution per input.

\clearpage

\printcredits

\section*{Declaration of competing interest}
Mustafa Fuad Rifet Ibrahim and Maurice Meijer are employed by NXP Semiconductors, whose MCXN947 microcontroller and integrated eIQ Neutron NPU are evaluated in this study. Alexander Schlaefer and Peer Stelldinger declare no competing interests.

\section*{Funding}
This work was supported by the Bundesministerium f\"ur Forschung, Technologie und Raumfahrt (BMFTR) [grant number 16KIS2451K (PATHCARE)]. The funder had no role in study design, data collection, analysis, interpretation, manuscript preparation, or the decision to submit the article for publication.

\section*{Ethics statement}
This study used existing, publicly available, de-identified physiological recordings from PTB-XL, BUT QDB, and the MIT-BIH Noise Stress Test Database. No new human-participant data were collected.

\section*{Data availability}
The physiological recordings are publicly available through PhysioNet~\cite{pollard2026physionet}: PTB-XL version 1.0.3~\cite{wagner2022ptbxl}, BUT QDB version 1.0.0~\cite{nemcova2020butqdb}, and the MIT-BIH Noise Stress Test Database version 1.0.0~\cite{moody1984noise}. The latter dataset is available at \url{https://doi.org/10.13026/C2HS3T}.
Aggregated numerical results underlying the reported figures and tables, together with processed hardware-measurement data, are available from the corresponding author on reasonable request. Raw current traces are no longer available.

\section*{Code availability}
Code for model training, architecture search, and experimental evaluation is available from the corresponding author on request during peer review and will be released in a public repository upon acceptance. Hardware-measurement code and embedded benchmark firmware are excluded from this release. The hardware measurement protocol is described in Section~\ref{subsec:measurement}.

\section*{Declaration of generative AI and AI-assisted technologies in the manuscript preparation process}
During the preparation of this work the authors used Anthropic's Claude Opus 5.5 and Microsoft 365 Copilot in order to assist with manuscript drafting and revision. After using this tool, the authors reviewed and edited the content as needed and take full responsibility for the content of the published article.

\bibliographystyle{pmc-num-names}

\bibliography{refs}

\begin{thebibliography}{57}
\expandafter\ifx\csname natexlab\endcsname\relax\def\natexlab#1{#1}\fi
\providecommand{\url}[1]{\texttt{#1}}
\providecommand{\href}[2]{#2}
\providecommand{\path}[1]{#1}
\providecommand{\DOIprefix}{doi:}
\providecommand{\ArXivprefix}{arXiv:}
\providecommand{\URLprefix}{URL: }
\providecommand{\Pubmedprefix}{pmid:}
\providecommand{\doi}[1]{\href{http://dx.doi.org/#1}{\path{#1}}}
\providecommand{\Pubmed}[1]{\href{pmid:#1}{\path{#1}}}
\providecommand{\bibinfo}[2]{#2}
\ifx\xfnm\relax \def\xfnm[#1]{\unskip,\space#1}\fi
%Type = Misc
\bibitem[{Ibrahim et~al.(2026)Ibrahim, Alkanat, Manthey, Meijer, Schlaefer, and
  Stelldinger}]{ibrahim2026ondevice}
\bibinfo{author}{M.~F.~R. Ibrahim}, \bibinfo{author}{T.~Alkanat},
  \bibinfo{author}{F.~Manthey}, \bibinfo{author}{M.~Meijer},
  \bibinfo{author}{A.~Schlaefer}, \bibinfo{author}{P.~Stelldinger},
  \bibinfo{title}{{On-device inference versus wireless streaming:
  Energy-efficient multi-modal deep learning for wearable cardiovascular
  patches}}, \bibinfo{year}{2026}. \DOIprefix\doi{10.48550/arXiv.2510.18668}.
  \href{http://arxiv.org/abs/2510.18668v5}{\tt arXiv:2510.18668v5},
  \bibinfo{note}{preprint}.
%Type = Article
\bibitem[{Venton et~al.(2021)Venton, Harris, Sundar, Smith, and
  Aston}]{venton2021robustness}
\bibinfo{author}{J.~Venton}, \bibinfo{author}{P.~M. Harris},
  \bibinfo{author}{A.~Sundar}, \bibinfo{author}{N.~A.~S. Smith},
  \bibinfo{author}{P.~J. Aston},
\newblock \bibinfo{title}{Robustness of convolutional neural networks to
  physiological electrocardiogram noise},
\newblock \bibinfo{journal}{Phil. Trans. R. Soc. A} \bibinfo{volume}{379}
  (\bibinfo{year}{2021}) \bibinfo{pages}{20200262}.
  \DOIprefix\doi{10.1098/rsta.2020.0262}.
%Type = Article
\bibitem[{Barandas et~al.(2024)Barandas, Famiglini, Campagner, Folgado,
  Sim{\~a}o, Cabitza, and Gamboa}]{barandas2024evaluation}
\bibinfo{author}{M.~Barandas}, \bibinfo{author}{L.~Famiglini},
  \bibinfo{author}{A.~Campagner}, \bibinfo{author}{D.~Folgado},
  \bibinfo{author}{R.~Sim{\~a}o}, \bibinfo{author}{F.~Cabitza},
  \bibinfo{author}{H.~Gamboa},
\newblock \bibinfo{title}{Evaluation of uncertainty quantification methods in
  multi-label classification: A case study with automatic diagnosis of
  electrocardiogram},
\newblock \bibinfo{journal}{Inf. Fusion} \bibinfo{volume}{101}
  (\bibinfo{year}{2024}) \bibinfo{pages}{101978}.
  \DOIprefix\doi{10.1016/j.inffus.2023.101978}.
%Type = Inproceedings
\bibitem[{Ovadia et~al.(2019)Ovadia, Fertig, Ren, Nado, Sculley, Nowozin,
  Dillon, Lakshminarayanan, and Snoek}]{ovadia2019can}
\bibinfo{author}{Y.~Ovadia}, \bibinfo{author}{E.~Fertig},
  \bibinfo{author}{J.~Ren}, \bibinfo{author}{Z.~Nado},
  \bibinfo{author}{D.~Sculley}, \bibinfo{author}{S.~Nowozin},
  \bibinfo{author}{J.~Dillon}, \bibinfo{author}{B.~Lakshminarayanan},
  \bibinfo{author}{J.~Snoek},
\newblock \bibinfo{title}{Can you trust your model's uncertainty? {Evaluating}
  predictive uncertainty under dataset shift},
\newblock in: \bibinfo{booktitle}{Advances in Neural Information Processing
  Systems}, volume~\bibinfo{volume}{32}, \bibinfo{publisher}{Curran Associates,
  Inc.}, \bibinfo{year}{2019}, pp. \bibinfo{pages}{13991--14002}. \URLprefix
  \url{https://proceedings.neurips.cc/paper/2019/hash/8558cb408c1d76621371888657d2eb1d-Abstract.html}.
%Type = Inproceedings
\bibitem[{Ibrahim et~al.(2026)Ibrahim, Meijer, Schlaefer, and
  Stelldinger}]{ibrahim2026enhancing}
\bibinfo{author}{M.~F.~R. Ibrahim}, \bibinfo{author}{M.~Meijer},
  \bibinfo{author}{A.~Schlaefer}, \bibinfo{author}{P.~Stelldinger},
\newblock \bibinfo{title}{Enhancing {ECG} classification robustness with
  lightweight unsupervised anomaly detection filters},
\newblock in: \bibinfo{booktitle}{2026 IEEE International Conference on
  Pervasive Computing and Communications Workshops and other Affiliated Events
  (PerCom Workshops)}, \bibinfo{organization}{IEEE}, \bibinfo{year}{2026}, pp.
  \bibinfo{pages}{1--6}.
  \DOIprefix\doi{10.1109/PerComWorkshops68308.2026.11585248}.
%Type = Inproceedings
\bibitem[{Gal and Ghahramani(2016)}]{gal2016dropout}
\bibinfo{author}{Y.~Gal}, \bibinfo{author}{Z.~Ghahramani},
\newblock \bibinfo{title}{Dropout as a {Bayesian} approximation: Representing
  model uncertainty in deep learning},
\newblock in: \bibinfo{booktitle}{Proceedings of The 33rd International
  Conference on Machine Learning}, volume~\bibinfo{volume}{48} of
  \textit{\bibinfo{series}{Proceedings of Machine Learning Research}},
  \bibinfo{publisher}{PMLR}, \bibinfo{year}{2016}, pp.
  \bibinfo{pages}{1050--1059}. \URLprefix
  \url{https://proceedings.mlr.press/v48/gal16.html}.
%Type = Inproceedings
\bibitem[{Lakshminarayanan et~al.(2017)Lakshminarayanan, Pritzel, and
  Blundell}]{lakshminarayanan2017simple}
\bibinfo{author}{B.~Lakshminarayanan}, \bibinfo{author}{A.~Pritzel},
  \bibinfo{author}{C.~Blundell},
\newblock \bibinfo{title}{Simple and scalable predictive uncertainty estimation
  using deep ensembles},
\newblock in: \bibinfo{booktitle}{Advances in Neural Information Processing
  Systems}, volume~\bibinfo{volume}{30}, \bibinfo{publisher}{Curran Associates,
  Inc.}, \bibinfo{year}{2017}, pp. \bibinfo{pages}{6402--6413}. \URLprefix
  \url{https://proceedings.neurips.cc/paper/2017/hash/9ef2ed4b7fd2c810847ffa5fa85bce38-Abstract.html}.
%Type = Inproceedings
\bibitem[{Atamny et~al.(2023)Atamny, Saguner, Abaecherli, and
  Konukoglu}]{atamny2023outlier}
\bibinfo{author}{O.~Atamny}, \bibinfo{author}{A.~Saguner},
  \bibinfo{author}{R.~Abaecherli}, \bibinfo{author}{E.~Konukoglu},
\newblock \bibinfo{title}{Outlier detection in {ECG}},
\newblock in: \bibinfo{booktitle}{2023 Computing in Cardiology (CinC)},
  volume~\bibinfo{volume}{50}, \bibinfo{organization}{IEEE},
  \bibinfo{year}{2023}, pp. \bibinfo{pages}{1--4}.
  \DOIprefix\doi{10.22489/CinC.2023.038}.
%Type = Article
\bibitem[{Zhou and Kan(2021)}]{zhou2021tensor}
\bibinfo{author}{H.~Zhou}, \bibinfo{author}{C.~Kan},
\newblock \bibinfo{title}{Tensor-based {ECG} anomaly detection toward cardiac
  monitoring in the internet of health things},
\newblock \bibinfo{journal}{Sensors} \bibinfo{volume}{21}
  (\bibinfo{year}{2021}) \bibinfo{pages}{4173}.
  \DOIprefix\doi{10.3390/s21124173}.
%Type = Inproceedings
\bibitem[{Jiang et~al.(2023)Jiang, Huang, Cao, Wu, Zeng, Chen, Zhang, and
  Wang}]{jiang2023multi}
\bibinfo{author}{A.~Jiang}, \bibinfo{author}{C.~Huang},
  \bibinfo{author}{Q.~Cao}, \bibinfo{author}{S.~Wu}, \bibinfo{author}{Z.~Zeng},
  \bibinfo{author}{K.~Chen}, \bibinfo{author}{Y.~Zhang},
  \bibinfo{author}{Y.~Wang},
\newblock \bibinfo{title}{Multi-scale cross-restoration framework for
  electrocardiogram anomaly detection},
\newblock in: \bibinfo{booktitle}{Medical Image Computing and Computer Assisted
  Intervention -- MICCAI 2023}, volume \bibinfo{volume}{14220} of
  \textit{\bibinfo{series}{Lecture Notes in Computer Science}},
  \bibinfo{publisher}{Springer}, \bibinfo{year}{2023}, pp.
  \bibinfo{pages}{87--97}. \DOIprefix\doi{10.1007/978-3-031-43907-0_9}.
%Type = Article
\bibitem[{Ruiz-Barroso et~al.(2025)Ruiz-Barroso, Castro, Miranda,
  Constantinescu, Atienza, and Guil}]{ruiz2025fade}
\bibinfo{author}{P.~Ruiz-Barroso}, \bibinfo{author}{F.~M. Castro},
  \bibinfo{author}{J.~Miranda}, \bibinfo{author}{D.-A. Constantinescu},
  \bibinfo{author}{D.~Atienza}, \bibinfo{author}{N.~Guil},
\newblock \bibinfo{title}{{FADE}: Forecasting for anomaly detection on {ECG}},
\newblock \bibinfo{journal}{Comput. Methods Programs Biomed.}
  \bibinfo{volume}{267} (\bibinfo{year}{2025}) \bibinfo{pages}{108780}.
  \DOIprefix\doi{10.1016/j.cmpb.2025.108780}.
%Type = Inproceedings
\bibitem[{Bui et~al.(2024)Bui, Hoang, Phan, Tran, Patel, Adjeroh, and
  Le}]{bui2024tsrnet}
\bibinfo{author}{N.-T. Bui}, \bibinfo{author}{D.-H. Hoang},
  \bibinfo{author}{T.~Phan}, \bibinfo{author}{M.-T. Tran},
  \bibinfo{author}{B.~Patel}, \bibinfo{author}{D.~Adjeroh},
  \bibinfo{author}{N.~Le},
\newblock \bibinfo{title}{{TSRNet}: Simple framework for real-time {ECG}
  anomaly detection with multimodal time and spectrogram restoration network},
\newblock in: \bibinfo{booktitle}{2024 IEEE International Symposium on
  Biomedical Imaging (ISBI)}, \bibinfo{organization}{IEEE},
  \bibinfo{year}{2024}, pp. \bibinfo{pages}{1--4}.
  \DOIprefix\doi{10.1109/ISBI56570.2024.10635676}.
%Type = Article
\bibitem[{Zhou et~al.(2026)Zhou, Yang, Gan, Li, Yuan, and
  Zhao}]{zhou2026lightweight}
\bibinfo{author}{Y.~Zhou}, \bibinfo{author}{Y.~Yang}, \bibinfo{author}{J.~Gan},
  \bibinfo{author}{X.~Li}, \bibinfo{author}{J.~Yuan},
  \bibinfo{author}{W.~Zhao},
\newblock \bibinfo{title}{A lightweight and robust method for electrocardiogram
  anomaly detection and localization using multi-scale masked autoencoder},
\newblock \bibinfo{journal}{PLOS ONE} \bibinfo{volume}{21}
  (\bibinfo{year}{2026}) \bibinfo{pages}{e0343571}.
  \DOIprefix\doi{10.1371/journal.pone.0343571}.
%Type = Misc
\bibitem[{Jiang et~al.(2024)Jiang, Huang, Cao, Xu, Zeng, Chen, Zhang, and
  Wang}]{jiang2024self}
\bibinfo{author}{A.~Jiang}, \bibinfo{author}{C.~Huang},
  \bibinfo{author}{Q.~Cao}, \bibinfo{author}{Y.~Xu}, \bibinfo{author}{Z.~Zeng},
  \bibinfo{author}{K.~Chen}, \bibinfo{author}{Y.~Zhang},
  \bibinfo{author}{Y.~Wang}, \bibinfo{title}{Self-supervised anomaly detection
  pretraining enhances long-tail {ECG} diagnosis}, \bibinfo{year}{2024}.
  \DOIprefix\doi{10.48550/arXiv.2408.17154}.
  \href{http://arxiv.org/abs/2408.17154}{\tt arXiv:2408.17154},
  \bibinfo{note}{preprint}.
%Type = Article
\bibitem[{Wang et~al.(2023)Wang, Stavrakis, and Yao}]{wang2023hierarchical}
\bibinfo{author}{Z.~Wang}, \bibinfo{author}{S.~Stavrakis},
  \bibinfo{author}{B.~Yao},
\newblock \bibinfo{title}{Hierarchical deep learning with generative
  adversarial network for automatic cardiac diagnosis from {ECG} signals},
\newblock \bibinfo{journal}{Comput. Biol. Med.} \bibinfo{volume}{155}
  (\bibinfo{year}{2023}) \bibinfo{pages}{106641}.
  \DOIprefix\doi{10.1016/j.compbiomed.2023.106641}.
%Type = Article
\bibitem[{Satija et~al.(2018)Satija, Ramkumar, and
  Manikandan}]{satija2017automated}
\bibinfo{author}{U.~Satija}, \bibinfo{author}{B.~Ramkumar},
  \bibinfo{author}{M.~S. Manikandan},
\newblock \bibinfo{title}{Automated {ECG} noise detection and classification
  system for unsupervised healthcare monitoring},
\newblock \bibinfo{journal}{IEEE J. Biomed. Health Inform.}
  \bibinfo{volume}{22} (\bibinfo{year}{2018}) \bibinfo{pages}{722--732}.
  \DOIprefix\doi{10.1109/JBHI.2017.2686436}.
%Type = Article
\bibitem[{Huang et~al.(2024)Huang, Zhang, Fan, Chang, Zhou, and
  Li}]{huang2024pseudo}
\bibinfo{author}{X.~Huang}, \bibinfo{author}{F.~Zhang},
  \bibinfo{author}{H.~Fan}, \bibinfo{author}{H.~Chang},
  \bibinfo{author}{B.~Zhou}, \bibinfo{author}{Z.~Li},
\newblock \bibinfo{title}{Pseudo anomalies enhanced deep support vector data
  description for electrocardiogram quality assessment},
\newblock \bibinfo{journal}{Comput. Biol. Med.} \bibinfo{volume}{170}
  (\bibinfo{year}{2024}) \bibinfo{pages}{107928}.
  \DOIprefix\doi{10.1016/j.compbiomed.2024.107928}.
%Type = Article
\bibitem[{Huang et~al.(2025)Huang, Xi, Fan, Zhang, Yu, Liu, Dong, and
  Guizani}]{huang2025joint}
\bibinfo{author}{X.~Huang}, \bibinfo{author}{L.~Xi}, \bibinfo{author}{H.~Fan},
  \bibinfo{author}{F.~Zhang}, \bibinfo{author}{X.~Yu},
  \bibinfo{author}{L.~Liu}, \bibinfo{author}{M.~Dong},
  \bibinfo{author}{M.~Guizani},
\newblock \bibinfo{title}{Joint time-frequency pseudo anomalies for multimodal
  electrocardiogram quality assessment in healthcare service computing},
\newblock \bibinfo{journal}{IEEE Trans. Serv. Comput.} \bibinfo{volume}{18}
  (\bibinfo{year}{2025}) \bibinfo{pages}{2158--2171}.
  \DOIprefix\doi{10.1109/TSC.2025.3583116}.
%Type = Article
\bibitem[{Fan et~al.(2026)Fan, Chang, Han, and Luo}]{fan2026global}
\bibinfo{author}{J.~Fan}, \bibinfo{author}{H.~Chang}, \bibinfo{author}{H.~Han},
  \bibinfo{author}{J.~Luo},
\newblock \bibinfo{title}{Global-key region collaborative noise-aware for
  electrocardiogram signal quality assessment},
\newblock \bibinfo{journal}{PLOS ONE} \bibinfo{volume}{21}
  (\bibinfo{year}{2026}) \bibinfo{pages}{e0355335}.
  \DOIprefix\doi{10.1371/journal.pone.0355335}.
%Type = Misc
\bibitem[{Han et~al.(2025)Han, Heo, Kim, Lee, Huh, Choi, Kim, and
  Kim}]{han2025diffusion}
\bibinfo{author}{T.-S. Han}, \bibinfo{author}{J.-W. Heo},
  \bibinfo{author}{H.~Kim}, \bibinfo{author}{C.-H. Lee},
  \bibinfo{author}{H.~Huh}, \bibinfo{author}{E.-K. Choi},
  \bibinfo{author}{H.~J. Kim}, \bibinfo{author}{D.-J. Kim},
  \bibinfo{title}{Diffusion-based electrocardiography noise quantification via
  anomaly detection}, \bibinfo{year}{2025}.
  \DOIprefix\doi{10.48550/arXiv.2506.11815}.
  \href{http://arxiv.org/abs/2506.11815}{\tt arXiv:2506.11815},
  \bibinfo{note}{preprint}.
%Type = Article
\bibitem[{Rahman et~al.(2025)Rahman, Pal, and Karmakar}]{rahman2025impact}
\bibinfo{author}{S.~Rahman}, \bibinfo{author}{S.~Pal},
  \bibinfo{author}{C.~Karmakar},
\newblock \bibinfo{title}{Impact and mitigation of blind filtering in automatic
  {ECG} analysis for wearable sensors},
\newblock \bibinfo{journal}{Biomed. Signal Process. Control}
  \bibinfo{volume}{110} (\bibinfo{year}{2025}) \bibinfo{pages}{108294}.
  \DOIprefix\doi{10.1016/j.bspc.2025.108294}.
%Type = Article
\bibitem[{Vranken et~al.(2021)Vranken, van~de Leur, Gupta, Juarez~Orozco,
  Hassink, van~der Harst, Doevendans, Gulshad, and van
  Es}]{vranken2021uncertainty}
\bibinfo{author}{J.~F. Vranken}, \bibinfo{author}{R.~R. van~de Leur},
  \bibinfo{author}{D.~K. Gupta}, \bibinfo{author}{L.~E. Juarez~Orozco},
  \bibinfo{author}{R.~J. Hassink}, \bibinfo{author}{P.~van~der Harst},
  \bibinfo{author}{P.~A. Doevendans}, \bibinfo{author}{S.~Gulshad},
  \bibinfo{author}{R.~van Es},
\newblock \bibinfo{title}{Uncertainty estimation for deep learning-based
  automated analysis of 12-lead electrocardiograms},
\newblock \bibinfo{journal}{Eur. Heart J. Digit. Health} \bibinfo{volume}{2}
  (\bibinfo{year}{2021}) \bibinfo{pages}{401--415}.
  \DOIprefix\doi{10.1093/ehjdh/ztab045}.
%Type = Article
\bibitem[{Elul et~al.(2021)Elul, Rosenberg, Schuster, Bronstein, and
  Yaniv}]{elul2021meeting}
\bibinfo{author}{Y.~Elul}, \bibinfo{author}{A.~A. Rosenberg},
  \bibinfo{author}{A.~Schuster}, \bibinfo{author}{A.~M. Bronstein},
  \bibinfo{author}{Y.~Yaniv},
\newblock \bibinfo{title}{Meeting the unmet needs of clinicians from {AI}
  systems showcased for cardiology with deep-learning--based {ECG} analysis},
\newblock \bibinfo{journal}{Proc. Natl. Acad. Sci. U.S.A.}
  \bibinfo{volume}{118} (\bibinfo{year}{2021}) \bibinfo{pages}{e2020620118}.
  \DOIprefix\doi{10.1073/pnas.2020620118}.
%Type = Article
\bibitem[{Yu et~al.(2025)Yu, Liu, Wu, Ren, and Zhao}]{yu2025trustworthy}
\bibinfo{author}{B.~Yu}, \bibinfo{author}{Y.~Liu}, \bibinfo{author}{X.~Wu},
  \bibinfo{author}{J.~Ren}, \bibinfo{author}{Z.~Zhao},
\newblock \bibinfo{title}{Trustworthy diagnosis of electrocardiography signals
  based on out-of-distribution detection},
\newblock \bibinfo{journal}{PLOS ONE} \bibinfo{volume}{20}
  (\bibinfo{year}{2025}) \bibinfo{pages}{e0317900}.
  \DOIprefix\doi{10.1371/journal.pone.0317900}.
%Type = Misc
\bibitem[{Shi et~al.(2025)Shi, Wang, and Li}]{shi2025unsupervised}
\bibinfo{author}{Z.~Shi}, \bibinfo{author}{Z.~Wang}, \bibinfo{author}{Y.~Li},
  \bibinfo{title}{Unsupervised identification and replay-based detection
  ({UIRD}) for new category anomaly detection in {ECG} signal},
  \bibinfo{year}{2025}. \DOIprefix\doi{10.48550/arXiv.2509.01512}.
  \href{http://arxiv.org/abs/2509.01512}{\tt arXiv:2509.01512},
  \bibinfo{note}{preprint}.
%Type = Article
\bibitem[{Rahman et~al.(2026)Rahman, Rivolta, Badilini, and
  Sassi}]{rahman2026uncertainty}
\bibinfo{author}{M.~M. Rahman}, \bibinfo{author}{M.~W. Rivolta},
  \bibinfo{author}{F.~Badilini}, \bibinfo{author}{R.~Sassi},
\newblock \bibinfo{title}{Uncertainty estimation of deep learning models for
  atrial fibrillation detection from {Holter} recordings: A benchmark study},
\newblock \bibinfo{journal}{Biomed. Signal Process. Control}
  \bibinfo{volume}{113} (\bibinfo{year}{2026}) \bibinfo{pages}{109032}.
  \DOIprefix\doi{10.1016/j.bspc.2025.109032}.
%Type = Inproceedings
\bibitem[{Jia et~al.(2024)Jia, Kwon, Ma, Pham, Qendro, Vu, and
  Mascolo}]{jia2024ur2m}
\bibinfo{author}{H.~Jia}, \bibinfo{author}{Y.~D. Kwon},
  \bibinfo{author}{D.~Ma}, \bibinfo{author}{N.~Pham},
  \bibinfo{author}{L.~Qendro}, \bibinfo{author}{T.~Vu},
  \bibinfo{author}{C.~Mascolo},
\newblock \bibinfo{title}{{UR2M}: Uncertainty and resource-aware event
  detection on microcontrollers},
\newblock in: \bibinfo{booktitle}{2024 IEEE International Conference on
  Pervasive Computing and Communications (PerCom)},
  \bibinfo{organization}{IEEE}, \bibinfo{year}{2024}, pp.
  \bibinfo{pages}{1--10}. \DOIprefix\doi{10.1109/PerCom59722.2024.10494467}.
%Type = Article
\bibitem[{Jia et~al.(2026)Jia, Kwon, Ma, Pham, Qendro, Vu, and
  Mascolo}]{jia2026cascade}
\bibinfo{author}{H.~Jia}, \bibinfo{author}{Y.~D. Kwon},
  \bibinfo{author}{D.~Ma}, \bibinfo{author}{N.~Pham},
  \bibinfo{author}{L.~Qendro}, \bibinfo{author}{T.~Vu},
  \bibinfo{author}{C.~Mascolo},
\newblock \bibinfo{title}{A cascade framework for on-device uncertainty-aware
  event detection on microcontrollers},
\newblock \bibinfo{journal}{Pervasive Mob. Comput.} \bibinfo{volume}{119}
  (\bibinfo{year}{2026}) \bibinfo{pages}{102208}.
  \DOIprefix\doi{10.1016/j.pmcj.2026.102208}.
%Type = Article
\bibitem[{Pollard et~al.(2026)Pollard, Moody, Lehman, Gow, Fernandes, Xie,
  Johnson, Mark, and Heldt}]{pollard2026physionet}
\bibinfo{author}{T.~Pollard}, \bibinfo{author}{B.~E. Moody},
  \bibinfo{author}{L.-w.~H. Lehman}, \bibinfo{author}{B.~J. Gow},
  \bibinfo{author}{C.~Fernandes}, \bibinfo{author}{C.~Xie},
  \bibinfo{author}{A.~Johnson}, \bibinfo{author}{R.~G. Mark},
  \bibinfo{author}{T.~Heldt},
\newblock \bibinfo{title}{{PhysioNet} as a global platform for biomedical
  research},
\newblock \bibinfo{journal}{Nat. Health} \bibinfo{volume}{1}
  (\bibinfo{year}{2026}) \bibinfo{pages}{792--795}.
  \DOIprefix\doi{10.1038/s44360-026-00096-z}.
%Type = Article
\bibitem[{Goldberger et~al.(2000)Goldberger, Amaral, Glass, Hausdorff, Ivanov,
  Mark, Mietus, Moody, Peng, and Stanley}]{goldberger2000physionet}
\bibinfo{author}{A.~L. Goldberger}, \bibinfo{author}{L.~A.~N. Amaral},
  \bibinfo{author}{L.~Glass}, \bibinfo{author}{J.~M. Hausdorff},
  \bibinfo{author}{P.~C. Ivanov}, \bibinfo{author}{R.~G. Mark},
  \bibinfo{author}{J.~E. Mietus}, \bibinfo{author}{G.~B. Moody},
  \bibinfo{author}{C.-K. Peng}, \bibinfo{author}{H.~E. Stanley},
\newblock \bibinfo{title}{{PhysioBank}, {PhysioToolkit}, and {PhysioNet}:
  Components of a new research resource for complex physiologic signals},
\newblock \bibinfo{journal}{Circulation} \bibinfo{volume}{101}
  (\bibinfo{year}{2000}) \bibinfo{pages}{e215--e220}.
  \DOIprefix\doi{10.1161/01.CIR.101.23.e215}.
%Type = Misc
\bibitem[{Wagner et~al.(2022)Wagner, Strodthoff, Bousseljot, Samek, and
  Schaeffter}]{wagner2022ptbxl}
[dataset] \bibinfo{author}{P.~Wagner}, \bibinfo{author}{N.~Strodthoff},
  \bibinfo{author}{R.-D. Bousseljot}, \bibinfo{author}{W.~Samek},
  \bibinfo{author}{T.~Schaeffter}, \bibinfo{title}{{PTB-XL}, a large publicly
  available electrocardiography dataset (version 1.0.3)},
  \bibinfo{howpublished}{PhysioNet}, \bibinfo{year}{2022}.
  \DOIprefix\doi{10.13026/kfzx-aw45}.
%Type = Article
\bibitem[{Wagner et~al.(2020)Wagner, Strodthoff, Bousseljot, Kreiseler, Lunze,
  Samek, and Schaeffter}]{wagner2020ptbxl}
\bibinfo{author}{P.~Wagner}, \bibinfo{author}{N.~Strodthoff},
  \bibinfo{author}{R.-D. Bousseljot}, \bibinfo{author}{D.~Kreiseler},
  \bibinfo{author}{F.~I. Lunze}, \bibinfo{author}{W.~Samek},
  \bibinfo{author}{T.~Schaeffter},
\newblock \bibinfo{title}{{PTB-XL}, a large publicly available
  electrocardiography dataset},
\newblock \bibinfo{journal}{Sci. Data} \bibinfo{volume}{7}
  (\bibinfo{year}{2020}) \bibinfo{pages}{154}.
  \DOIprefix\doi{10.1038/s41597-020-0495-6}.
%Type = Misc
\bibitem[{Nemcova et~al.(2020)Nemcova, Smisek, Opravilov{\'a}, Vitek, Smital,
  and Mar{\v s}{\'a}nov{\'a}}]{nemcova2020butqdb}
[dataset] \bibinfo{author}{A.~Nemcova}, \bibinfo{author}{R.~Smisek},
  \bibinfo{author}{K.~Opravilov{\'a}}, \bibinfo{author}{M.~Vitek},
  \bibinfo{author}{L.~Smital}, \bibinfo{author}{L.~Mar{\v s}{\'a}nov{\'a}},
  \bibinfo{title}{{Brno University of Technology ECG Quality Database (BUT
  QDB)} (version 1.0.0)}, \bibinfo{howpublished}{PhysioNet},
  \bibinfo{year}{2020}. \DOIprefix\doi{10.13026/kah4-0w24}.
%Type = Inproceedings
\bibitem[{Moody et~al.(1984)Moody, Muldrow, and Mark}]{moody1984noise}
\bibinfo{author}{G.~B. Moody}, \bibinfo{author}{W.~K. Muldrow},
  \bibinfo{author}{R.~G. Mark},
\newblock \bibinfo{title}{{A noise stress test for arrhythmia detectors}},
\newblock in: \bibinfo{booktitle}{Computers in Cardiology},
  volume~\bibinfo{volume}{11}, \bibinfo{year}{1984}, pp.
  \bibinfo{pages}{381--384}.
%Type = Misc
\bibitem[{Moody(2022)}]{moody2022nst}
\bibinfo{author}{G.~B. Moody}, \bibinfo{title}{{nst -- noise stress test for
  ECG analysis programs}},
  \bibinfo{howpublished}{\url{https://physionet.org/physiotools/wag/nst-1.htm}},
  \bibinfo{year}{2022}. \bibinfo{note}{Accessed: 2025-08-25}.
%Type = Article
\bibitem[{Donoho(1995)}]{donoho1995denoising}
\bibinfo{author}{D.~L. Donoho},
\newblock \bibinfo{title}{{De-noising by soft-thresholding}},
\newblock \bibinfo{journal}{IEEE Trans. Inf. Theory} \bibinfo{volume}{41}
  (\bibinfo{year}{1995}) \bibinfo{pages}{613--627}.
  \DOIprefix\doi{10.1109/18.382009}.
%Type = Inproceedings
\bibitem[{Ruff et~al.(2018)Ruff, Vandermeulen, Goernitz, Deecke, Siddiqui,
  Binder, M{\"u}ller, and Kloft}]{ruff2018deep}
\bibinfo{author}{L.~Ruff}, \bibinfo{author}{R.~Vandermeulen},
  \bibinfo{author}{N.~Goernitz}, \bibinfo{author}{L.~Deecke},
  \bibinfo{author}{S.~A. Siddiqui}, \bibinfo{author}{A.~Binder},
  \bibinfo{author}{E.~M{\"u}ller}, \bibinfo{author}{M.~Kloft},
\newblock \bibinfo{title}{{Deep one-class classification}},
\newblock in: \bibinfo{booktitle}{Proceedings of the 35th International
  Conference on Machine Learning}, volume~\bibinfo{volume}{80} of
  \textit{\bibinfo{series}{Proceedings of Machine Learning Research}},
  \bibinfo{publisher}{PMLR}, \bibinfo{year}{2018}, pp.
  \bibinfo{pages}{4393--4402}. \URLprefix
  \url{https://proceedings.mlr.press/v80/ruff18a.html}.
%Type = Article
\bibitem[{Hinton and Salakhutdinov(2006)}]{hinton2006reducing}
\bibinfo{author}{G.~E. Hinton}, \bibinfo{author}{R.~R. Salakhutdinov},
\newblock \bibinfo{title}{Reducing the dimensionality of data with neural
  networks},
\newblock \bibinfo{journal}{Science} \bibinfo{volume}{313}
  (\bibinfo{year}{2006}) \bibinfo{pages}{504--507}.
  \DOIprefix\doi{10.1126/science.1127647}.
%Type = Inproceedings
\bibitem[{Kingma and Welling(2014)}]{kingma2013auto}
\bibinfo{author}{D.~P. Kingma}, \bibinfo{author}{M.~Welling},
\newblock \bibinfo{title}{Auto-encoding variational {Bayes}},
\newblock in: \bibinfo{booktitle}{International Conference on Learning
  Representations}, \bibinfo{year}{2014}. \URLprefix
  \url{https://arxiv.org/abs/1312.6114}.
%Type = Inproceedings
\bibitem[{Fu and Xue(2022)}]{fu2022mad}
\bibinfo{author}{Y.~Fu}, \bibinfo{author}{F.~Xue},
\newblock \bibinfo{title}{{MAD}: Self-supervised masked anomaly detection task
  for multivariate time series},
\newblock in: \bibinfo{booktitle}{2022 International Joint Conference on Neural
  Networks (IJCNN)}, \bibinfo{organization}{IEEE}, \bibinfo{year}{2022}, pp.
  \bibinfo{pages}{1--8}. \DOIprefix\doi{10.1109/IJCNN55064.2022.9892218}.
%Type = Inproceedings
\bibitem[{Kingma and Dhariwal(2018)}]{kingma2018glow}
\bibinfo{author}{D.~P. Kingma}, \bibinfo{author}{P.~Dhariwal},
\newblock \bibinfo{title}{{Glow: Generative flow with invertible 1x1
  convolutions}},
\newblock in: \bibinfo{booktitle}{Advances in Neural Information Processing
  Systems}, volume~\bibinfo{volume}{31}, \bibinfo{publisher}{Curran Associates,
  Inc.}, \bibinfo{year}{2018}, pp. \bibinfo{pages}{10215--10224}. \URLprefix
  \url{https://proceedings.neurips.cc/paper_files/paper/2018/hash/d139db6a236200b21cc7f752979132d0-Abstract.html}.
%Type = Inproceedings
\bibitem[{Ho et~al.(2020)Ho, Jain, and Abbeel}]{ho2020denoising}
\bibinfo{author}{J.~Ho}, \bibinfo{author}{A.~Jain},
  \bibinfo{author}{P.~Abbeel},
\newblock \bibinfo{title}{Denoising diffusion probabilistic models},
\newblock in: \bibinfo{booktitle}{Advances in Neural Information Processing
  Systems}, volume~\bibinfo{volume}{33}, \bibinfo{publisher}{Curran Associates,
  Inc.}, \bibinfo{year}{2020}, pp. \bibinfo{pages}{6840--6851}. \URLprefix
  \url{https://proceedings.neurips.cc/paper/2020/hash/4c5bcfec8584af0d967f1ab10179ca4b-Abstract.html}.
%Type = Inproceedings
\bibitem[{Wyatt et~al.(2022)Wyatt, Leach, Schmon, and
  Willcocks}]{wyatt2022anoddpm}
\bibinfo{author}{J.~Wyatt}, \bibinfo{author}{A.~Leach}, \bibinfo{author}{S.~M.
  Schmon}, \bibinfo{author}{C.~G. Willcocks},
\newblock \bibinfo{title}{{AnoDDPM}: Anomaly detection with denoising diffusion
  probabilistic models using simplex noise},
\newblock in: \bibinfo{booktitle}{{2022 IEEE/CVF Conference on Computer Vision
  and Pattern Recognition Workshops (CVPRW)}}, \bibinfo{organization}{IEEE},
  \bibinfo{year}{2022}, pp. \bibinfo{pages}{649--655}.
  \DOIprefix\doi{10.1109/CVPRW56347.2022.00080}.
%Type = Article
\bibitem[{Strodthoff et~al.(2021)Strodthoff, Wagner, Schaeffter, and
  Samek}]{strodthoff2020deep}
\bibinfo{author}{N.~Strodthoff}, \bibinfo{author}{P.~Wagner},
  \bibinfo{author}{T.~Schaeffter}, \bibinfo{author}{W.~Samek},
\newblock \bibinfo{title}{Deep learning for {ECG} analysis: Benchmarks and
  insights from {PTB-XL}},
\newblock \bibinfo{journal}{IEEE J. Biomed. Health Inform.}
  \bibinfo{volume}{25} (\bibinfo{year}{2021}) \bibinfo{pages}{1519--1528}.
  \DOIprefix\doi{10.1109/JBHI.2020.3022989}.
%Type = Inproceedings
\bibitem[{Gal and Ghahramani(2016)}]{gal2016bayesianconv}
\bibinfo{author}{Y.~Gal}, \bibinfo{author}{Z.~Ghahramani},
\newblock \bibinfo{title}{{Bayesian} convolutional neural networks with
  {Bernoulli} approximate variational inference},
\newblock in: \bibinfo{booktitle}{International Conference on Learning
  Representations, Workshop Track}, \bibinfo{year}{2016}. \URLprefix
  \url{https://openreview.net/pdf?id=3QxqXoJEyfp7y9wltP11}.
%Type = Misc
\bibitem[{Houlsby et~al.(2011)Houlsby, Husz{\'a}r, Ghahramani, and
  Lengyel}]{houlsby2011bayesian}
\bibinfo{author}{N.~Houlsby}, \bibinfo{author}{F.~Husz{\'a}r},
  \bibinfo{author}{Z.~Ghahramani}, \bibinfo{author}{M.~Lengyel},
  \bibinfo{title}{{Bayesian} active learning for classification and preference
  learning}, \bibinfo{year}{2011}. \DOIprefix\doi{10.48550/arXiv.1112.5745}.
  \href{http://arxiv.org/abs/1112.5745}{\tt arXiv:1112.5745},
  \bibinfo{note}{preprint}.
%Type = Misc
\bibitem[{{NXP Semiconductors}(nd)}]{nxpNeutronConverter}
\bibinfo{author}{{NXP Semiconductors}}, \bibinfo{title}{{Neutron Converter}},
  \bibinfo{howpublished}{eIQ Learning Hub}, \bibinfo{year}{n.d.} \URLprefix
  \url{https://eiq.nxp.com/learning-hub/tools/neutronSdk/softwareTools/NeutronConverter.html},
  \bibinfo{note}{accessed: 2026-09-21}.
%Type = Inproceedings
\bibitem[{David et~al.(2021)David, Duke, Jain, Janapa~Reddi, Jeffries, Li,
  Kreeger, Nappier, Natraj, Wang, Warden, and Rhodes}]{david2021tensorflow}
\bibinfo{author}{R.~David}, \bibinfo{author}{J.~Duke},
  \bibinfo{author}{A.~Jain}, \bibinfo{author}{V.~Janapa~Reddi},
  \bibinfo{author}{N.~Jeffries}, \bibinfo{author}{J.~Li},
  \bibinfo{author}{N.~Kreeger}, \bibinfo{author}{I.~Nappier},
  \bibinfo{author}{M.~Natraj}, \bibinfo{author}{T.~Wang},
  \bibinfo{author}{P.~Warden}, \bibinfo{author}{R.~Rhodes},
\newblock \bibinfo{title}{{TensorFlow Lite Micro}: Embedded machine learning
  for {TinyML} systems},
\newblock in: \bibinfo{booktitle}{Proceedings of Machine Learning and Systems},
  volume~\bibinfo{volume}{3}, \bibinfo{year}{2021}, pp.
  \bibinfo{pages}{800--811}. \URLprefix
  \url{https://proceedings.mlsys.org/paper_files/paper/2021/hash/6c44dc73014d66ba49b28d483a8f8b0d-Abstract.html}.
%Type = Manual
\bibitem[{{NXP Semiconductors}(2025)}]{nxp2025mcxn947}
\bibinfo{author}{{NXP Semiconductors}},
  \bibinfo{title}{{MCXN947/946/547/546/537/536/527/526/247}: Highly integrated
  low-power dual-core {Arm Cortex-M33 MCU} with on-chip accelerators},
  \bibinfo{organization}{NXP Semiconductors}, \bibinfo{year}{2025}. \URLprefix
  \url{https://www.nxp.com/docs/en/data-sheet/MCXNP184M150F70.pdf},
  \bibinfo{note}{product data sheet, Rev. 8, 28 October 2025}.
%Type = Misc
\bibitem[{Bamberg et~al.(2025)Bamberg, Minnella, Bosio, Ottati, Wang, Lee,
  Lavagno, and Fuks}]{bamberg2025eiq}
\bibinfo{author}{L.~Bamberg}, \bibinfo{author}{F.~Minnella},
  \bibinfo{author}{R.~Bosio}, \bibinfo{author}{F.~Ottati},
  \bibinfo{author}{Y.~Wang}, \bibinfo{author}{J.~Lee},
  \bibinfo{author}{L.~Lavagno}, \bibinfo{author}{A.~Fuks}, \bibinfo{title}{{eIQ
  Neutron}: Redefining edge-{AI} inference with integrated {NPU} and compiler
  innovations}, \bibinfo{year}{2025}.
  \DOIprefix\doi{10.48550/arXiv.2509.14388}.
  \href{http://arxiv.org/abs/2509.14388}{\tt arXiv:2509.14388},
  \bibinfo{note}{preprint}.
%Type = Manual
\bibitem[{{NXP Semiconductors}(2023)}]{nxp2023mculink}
\bibinfo{author}{{NXP Semiconductors}}, \bibinfo{title}{{MCU-Link Pro}},
  \bibinfo{organization}{NXP Semiconductors}, \bibinfo{year}{2023}. \URLprefix
  \url{https://www.nxp.com/docs/en/user-manual/UM11673.pdf},
  \bibinfo{note}{user manual UM11673, Rev. 1.2, 26 June 2023}.
%Type = Inproceedings
\bibitem[{Shin et~al.(2013)Shin, Park, Kim, Lee, Ko, and Kim}]{shin2013ultra}
\bibinfo{author}{K.~Shin}, \bibinfo{author}{G.~G. Park}, \bibinfo{author}{J.~P.
  Kim}, \bibinfo{author}{T.~H. Lee}, \bibinfo{author}{B.~H. Ko},
  \bibinfo{author}{Y.~H. Kim},
\newblock \bibinfo{title}{An ultra-low power ({ULP}) bandage-type {ECG} sensor
  for efficient cardiac disease management},
\newblock in: \bibinfo{booktitle}{2013 35th Annual International Conference of
  the IEEE Engineering in Medicine and Biology Society (EMBC)},
  \bibinfo{publisher}{IEEE}, \bibinfo{year}{2013}, pp.
  \bibinfo{pages}{1474--1477}. \DOIprefix\doi{10.1109/EMBC.2013.6609790}.
%Type = Article
\bibitem[{Etemadi et~al.(2016)Etemadi, Inan, Heller, Hersek, Klein, and
  Roy}]{etemadi2016wearable}
\bibinfo{author}{M.~Etemadi}, \bibinfo{author}{O.~T. Inan},
  \bibinfo{author}{J.~A. Heller}, \bibinfo{author}{S.~Hersek},
  \bibinfo{author}{L.~Klein}, \bibinfo{author}{S.~Roy},
\newblock \bibinfo{title}{A wearable patch to enable long-term monitoring of
  environmental, activity and hemodynamics variables},
\newblock \bibinfo{journal}{IEEE Trans. Biomed. Circuits Syst.}
  \bibinfo{volume}{10} (\bibinfo{year}{2016}) \bibinfo{pages}{280--288}.
  \DOIprefix\doi{10.1109/TBCAS.2015.2405480}.
%Type = Article
\bibitem[{Hancock et~al.(2009)Hancock, Deal, Mirvis, Okin, Kligfield, and
  Gettes}]{hancock2009aha}
\bibinfo{author}{E.~W. Hancock}, \bibinfo{author}{B.~J. Deal},
  \bibinfo{author}{D.~M. Mirvis}, \bibinfo{author}{P.~Okin},
  \bibinfo{author}{P.~Kligfield}, \bibinfo{author}{L.~S. Gettes},
\newblock \bibinfo{title}{{AHA/ACCF/HRS} recommendations for the
  standardization and interpretation of the electrocardiogram: Part {V}:
  Electrocardiogram changes associated with cardiac chamber hypertrophy: A
  scientific statement from the {American Heart Association Electrocardiography
  and Arrhythmias Committee, Council on Clinical Cardiology}; the {American
  College of Cardiology Foundation}; and the {Heart Rhythm Society}. {Endorsed}
  by the {International Society for Computerized Electrocardiology}},
\newblock \bibinfo{journal}{J. Am. Coll. Cardiol.} \bibinfo{volume}{53}
  (\bibinfo{year}{2009}) \bibinfo{pages}{992--1002}.
  \DOIprefix\doi{10.1016/j.jacc.2008.12.015}.
%Type = Misc
\bibitem[{Bouman and Heskes(2025)}]{bouman2025autoencoders}
\bibinfo{author}{R.~Bouman}, \bibinfo{author}{T.~Heskes},
  \bibinfo{title}{{Autoencoders for anomaly detection are unreliable}},
  \bibinfo{year}{2025}. \DOIprefix\doi{10.48550/arXiv.2501.13864}.
  \href{http://arxiv.org/abs/2501.13864}{\tt arXiv:2501.13864},
  \bibinfo{note}{preprint}.
%Type = Inproceedings
\bibitem[{Kirichenko et~al.(2020)Kirichenko, Izmailov, and
  Wilson}]{kirichenko2020why}
\bibinfo{author}{P.~Kirichenko}, \bibinfo{author}{P.~Izmailov},
  \bibinfo{author}{A.~G. Wilson},
\newblock \bibinfo{title}{{Why normalizing flows fail to detect
  out-of-distribution data}},
\newblock in: \bibinfo{booktitle}{Advances in Neural Information Processing
  Systems}, volume~\bibinfo{volume}{33}, \bibinfo{publisher}{Curran Associates,
  Inc.}, \bibinfo{year}{2020}, pp. \bibinfo{pages}{20578--20589}. \URLprefix
  \url{https://proceedings.neurips.cc/paper/2020/hash/ecb9fe2fbb99c31f567e9823e884dbec-Abstract.html}.
%Type = Inproceedings
\bibitem[{Dehghani et~al.(2022)Dehghani, Arnab, Beyer, Vaswani, and
  Tay}]{dehghani2022efficiency}
\bibinfo{author}{M.~Dehghani}, \bibinfo{author}{A.~Arnab},
  \bibinfo{author}{L.~Beyer}, \bibinfo{author}{A.~Vaswani},
  \bibinfo{author}{Y.~Tay},
\newblock \bibinfo{title}{The efficiency misnomer},
\newblock in: \bibinfo{booktitle}{International Conference on Learning
  Representations}, \bibinfo{publisher}{OpenReview.net}, \bibinfo{year}{2022}.
  \URLprefix \url{https://openreview.net/forum?id=iulEMLYh1uR}.

\end{thebibliography}

\end{document}